\documentclass{article}

\PassOptionsToPackage{numbers}{natbib}

\usepackage[preprint]{neurips_2020}

\usepackage[utf8]{inputenc}
\usepackage[T1]{fontenc}
\usepackage{hyperref}
\usepackage{url}
\usepackage{booktabs}
\usepackage{amsfonts}
\usepackage{nicefrac}
\usepackage{microtype}

\usepackage{upgreek}
\usepackage{amsmath,bm}
\usepackage{bbm}
\usepackage{algorithm,algorithmic}
\usepackage{graphicx,wrapfig}
\usepackage{caption}
\usepackage{subcaption}

\newtheorem{theorem}{Theorem}
\newcommand*{\qedsquare}{\hfill\ensuremath{\square}}

\DeclareMathOperator{\measureP}{\mathbb{P}}
\DeclareMathOperator{\measureE}{\mathbb{E}}
\newcommand{\prob}[1]{\measureP{\left(#1\right)}}
\newcommand{\probL}[2]{\measureP_{#1}{\left(#2\right)}}

\newcommand{\EL}[2]{\measureE_{#1}{\left[#2\right]}}

\allowdisplaybreaks

\title{Contextual Policy Transfer in Reinforcement Learning Domains via Deep Mixtures-of-Experts}

\author{
  Michael Gimelfarb\thanks{Affiliate to Vector Institute of Artificial Intelligence, Toronto}\\
  Mechanical and Industrial Engineering \\ University of Toronto\\
  \texttt{mike.gimelfarb@mail.utoronto.ca}\\
  \And
   Scott Sanner \\
   Mechanical and Industrial Engineering \\ University of Toronto\\
   \texttt{ssanner@mie.utoronto.ca}\\
  \And
   Chi-Guhn Lee \\
   Mechanical and Industrial Engineering \\ University of Toronto\\  
   \texttt{cglee@mie.utoronto.ca}
}

\begin{document}

\maketitle

\begin{abstract}
    In reinforcement learning, agents that consider the context, or current state, when selecting source policies for transfer have been shown to outperform context-free approaches. However, none of the existing approaches transfer knowledge contextually from model-based learners to a model-free learner. This could be useful, for instance, when source policies are intentionally learned on diverse simulations with plentiful data but transferred to a real-world setting with limited data. In this paper, we assume knowledge of estimated source task dynamics and policies, and common sub-goals but different dynamics. We introduce a novel deep mixture-of-experts formulation for learning state-dependent beliefs over source task dynamics that match the target dynamics using state trajectories collected from the target task. The mixture model is easy to interpret, demonstrates robustness to estimation errors in dynamics, and is compatible with most learning algorithms. We then show how this model can be incorporated into standard policy reuse frameworks, and demonstrate its effectiveness on benchmarks from OpenAI-Gym.
\end{abstract}

\section{Introduction}

\emph{Reinforcement learning} (RL) is a general framework for the development of artificial agents that learn to make complex decisions by interacting with an environment. In recent years, off-the-shelf RL algorithms have achieved state-of-the-art performance on simulated tasks such as Atari games \citep{mnih2015human} and some real-world applications \citep{gu2017deep}. However, model-free RL algorithms are sensitive to the choice of reward function or hyper-parameters \citep{henderson2018deep}, and often require many samples to converge to good policies without prior knowledge \citep{yarats2019improving}.

To address this concern, \emph{transfer learning} tries to reduce the number of samples required to learn a new (target) task by reusing previously acquired knowledge from other similar (source) tasks \citep{lazaric2012transfer,taylor2009transfer}. Many papers in this area focus on reusing policies \citep{barreto2017successor,brys2015policy,gupta2017,parisotto2015actor} because it is intuitive, direct, and does not rely on value functions that can be difficult to transfer or not available. Furthermore, transferring policies from multiple source tasks can be more effective than a single policy \citep{comanici2010optimal,fernandez2006probabilistic,rosman2016bayesian}. However, despite recent developments in transfer learning theory, 
few existing approaches are able to reason \emph{contextually} about task similarity in different regions of the state space. This could lead to significant improvements in policy transfer, since only the information from relevant regions of each source task can be selected for transfer in a mix-and-match manner~\citep{taylor2009transfer}. 

In this paper, we assume that sub-goals are shared between source and target tasks, but dynamics can vary \citep{konidaris2007building,tirinzoni2019transfer}. Furthermore, the dynamics and control policies of the source tasks are estimated prior to transfer. Such knowledge is often naturally available in operations research settings, including queuing systems, or inventory and supply chain management problems, where the dynamics models can be specified in advance but depend on key parameters that need to be learned, such as the demand distribution \citep{azoury1985bayes}.  In more general settings, it can be advantageous to transfer from a simulator to the real world \citep{tan2018sim}, or use model-free RL to fine-tune a policy obtained using a model-based learner \citep{nagabandi2018neural}. Furthermore, we do not require that the source dynamics be learned precisely for good transfer, as demonstrated in our experiments. 

This problem formulation has many potential applications in practice. In asset maintenance, practitioners often rely on a digital reconstruction of the machine and the environment, called a \emph{digital twin}, to learn a repair and replacement policy \citep{el2018digital,lund2018digital}. Here, source tasks could represent models of optimal control under a wide range of conditions, including weather patterns and degradation properties and usage of the machine, corresponding to common state, action, and goals, but differing dynamics. In manufacturing and supply chain management, when a new good is produced and sold, the dynamics of the stock level (state) might depend on the unknown demand distribution for the new product \citep{azoury1985bayes}. In this case, analysts can produce different projections for demand based on similar existing products or surveys, and manufacturing or stock replenishment policies can be learned under these different scenarios and transferred. Simulations such as these are routinely developed in many fields, such as drug discovery \citep{durrant2011molecular}, robotics \citep{christiano2016transfer} or manufacturing \citep{zhang2017digital}. Furthermore, in order to achieve robustness in these settings, one can simulate high-risk scenarios that do not occur often in practice and contextually transfer this information to the real-world target learner to enhance its safety during unexpected events.

To enable contextual policy transfer in this setting, we introduce a novel Bayesian framework for autonomously identifying and combining promising sub-regions from multiple source tasks. This is done by placing state-dependent Dirichlet priors over source task models, and updating them using state trajectories sampled from the target dynamics while exploring the target task. Specifically, posterior updates are informed by the likelihood of the observed transitions under the source task models. However, explicit knowledge of target dynamics is not necessary, making our approach model-free with respect to the target task, which is a critical assumption when the target task has sparse data.
Furthermore, naive tabulation of state-dependent priors is intractable in large or continuous state spaces, so we parameterize them as deep neural networks. Interpreted as a contextual \emph{mixture-of-experts} (MoE) \citep{bishop1994mixture,jacobs1991adaptive}, this architecture serves as a surrogate model for informing the state-dependent contextual selection of source policies for locally exploring promising actions in each state.

Our approach has several key advantages over other existing methods. Firstly, Bayesian inference allows priors to be specified over source task models, and is robust to noisy or imprecise dynamics. Secondly, the mixture model network can benefit from advances in (deep) function approximation~\citep{krizhevsky2012imagenet}, while enabling contextual transfer in a Bayesian framework.
Finally, our approach separates reasoning about task similarity from policy learning, making it agnostic to the reinforcement learning algorithm and the type of knowledge transferred. It is also easy to interpret, as we demonstrate later in our experiments (Figure~\ref{fig:learning_representation}). 

The main contributions of this paper are threefold:
\begin{enumerate}
    \item We introduce a \emph{contextual} mixture-of-experts model to efficiently learn state-dependent posterior distributions over source task models;
    \item We show how the trained mixture model can be incorporated into existing policy reuse methods, such as policy shaping (MAPSE) and reward shaping (MARS);
    \item We demonstrate the effectiveness and generality of our approach by testing it on problems with discrete and continuous spaces, including physics simulations.
\end{enumerate}

\section{Preliminaries}
\label{sec:prelim}

\paragraph{Markov Decision Process}
We follow the framework of \emph{Markov decision processes} (MDPs) \citep{puterman2014markov}, defined as five-tuples $\langle \mathcal{S}, \mathcal{A}, P, R, \gamma \rangle$ where: $\mathcal{S}$ is a set of states, $\mathcal{A}$ is a set of actions, $P(s' | s, a)$ are the state dynamics, $R(s, a, s')$ is a bounded reward function, and $\gamma \in [0, 1]$ is a discount factor. In deterministic problems, the state dynamics are typically represented as a deterministic map $s' = f(s,a)$. The objective of an agent is to find an optimal deterministic policy $\pi^*$ that maximizes the discounted cumulative reward $
	Q^\pi(s, a) = \EL{s \sim P, a \sim \pi}{r_0 + \gamma r_1 + \,\dots\, + \gamma^T r_T {|} s_0 = s, a_0 = a}$
over the planning horizon $t \in [0, T]$, where $r_t = R(s_t, a_t, s_{t+1})$.

\paragraph{Reinforcement Learning} 
In the reinforcement learning setting, neither $P$ nor $R$ are assumed to be known by the agent a-priori. Instead, an agent collects data $(s_0, a_0, r_1, s_1, a_1, r_1, \dots s_T)$ by interacting with the environment through a randomized exploration policy. In \emph{model-based RL} (MBRL), the agent uses this data to first learn $P$ and $R$, and then uses this to learn the optimal policy $\pi^*$. In \emph{model-free RL}, an agent learns the optimal policy $\pi^*$ directly without knowledge of $P$ or $R$. Model-free RL algorithms either approximate $Q^\pi$ \citep{mnih2015human,watkins1992q}, or parameterize and learn the optimal policy directly \citep{sutton2000policy}.

\paragraph{Learning Dynamics} 
In model-based RL, the dynamics model $f(s,a)$ or $P(s'|s,a)$ is typically parameterized as a deep neural network and trained through repeated interaction with the environment. It returns an estimate of the next state directly $\hat{s}' = \hat{f}_{\bm{\phi}}(s,a)$, or approximates its distribution, for instance using a Gaussian model $\hat{s}' \sim \mathcal{N}\left(\bm{\mu}_{\bm{\phi}}(s,a), \bm{\Sigma}_{\bm{\phi}}(s,a)\right)$. Subsequently, samples from the trained dynamics model can be used to augment the real experience when training the policy \citep{kaiser2019model,peng2018deep,sutton1991dyna}, or the dynamics can be used in other ways \citep{heess2015learning,levine2013guided,nagabandi2018neural,todorov2005generalized}. In this paper, we use dynamics to do efficient and robust transfer of knowledge between tasks.

\paragraph{Transfer Learning} 
We are interested in solving the following transfer learning problem. A library of $n \geq 1$ source tasks, and a single target task, are provided with identical $\mathcal{S}$ and $\mathcal{A}$ and shared sub-goals, but different dynamics. For each source task $i = 1, 2 \dots n$, the control policy $\pi_i^*$ and the underlying dynamics function $\hat{P}_i(s'|s,a)$ or $\hat{f}_{i}(s,a)$ is estimated. More generally, it is possible to transfer sample data, value functions, or other sources of domain knowledge in our framework, but we only study policy transfer in this paper. The main objective is to make use of this knowledge to solve the new target task in an efficient online manner.

\section{Model-Aware Policy Transfer}
\label{sec:theory}

In many domains, the state dynamics of a target task may be locally similar to one source task in one region of the state space, but a different source task in another region. By reasoning about task similarity \emph{locally} in different regions of the state space, an RL agent can make more efficient use of source task knowledge \citep{taylor2007transfer}. In this section, we proceed to model and learn state-dependent contextual similarity between source tasks and a target task. We also derive a theoretical result to justify our use of source task dynamics.

\subsection{Deep Contextual Mixture-of-Experts}
\label{subsec:bmc}

We first introduce a state-dependent prior $\prob{\mathbf{w} | s, \mathcal{D}_t}$ over {combinations} $\mathbf{w}$ of source task models, that tries to match the true (unknown) target dynamics using transitions $\mathcal{D}_t = \lbrace (s_\tau, a_\tau, s_{\tau+1}), \,\tau=1,2\dots t\rbrace$ collected from the target environment up to current time $t$. Here, $\mathbf{w} \in \mathbb{R}^n$ consists of non-negative elements such that $\sum_{i=1}^n w_i = 1$. Using \emph{combinations} to model uncertainty in source task selection can be viewed as \emph{Bayesian model combination}, 
exhibits stable convergence, is more robust to model misspecification, and allows inference over mixtures of models \citep{minka2000bayesian,monteith2011turning}.

In this setting, exact inference over $\mathbf{w}$ is intractable, so we model $\mathbf{w}$ using a surrogate probability distribution. In particular, since each realization of $\mathbf{w}$ is a discrete probability distribution, a suitable posterior for $\mathbf{w}$ in each state $s$ is a Dirichlet distribution with density $\prob{\mathbf{w}|s, \mathcal{D}_t} \propto \prod_{i=1}^n w_i^{\alpha_{t,i}(s)-1}$,
where $\alpha_{t,i} : \mathcal{S} \to \mathbb{R}$ for all $i =1, 2 \dots n$ and $t = 0, 1, 2, \dots$ and is non-negative. By averaging out the uncertainty in $\mathbf{w}$, we can obtain a posterior estimator $\prob{s'|s,a,\mathcal{D}_t}$ of target dynamics:
\begin{align}
    \prob{s'|s,a,\mathcal{D}_t} &= \int \prob{s'|s,a,\mathbf{w}}\prob{\mathbf{w}|s,\mathcal{D}_t} \,\mathrm{d}\mathbf{w} = \int \sum_{i=1}^n \prob{s'|s,a,\mathbf{w}, i} \prob{i|\mathbf{w},s} \prob{\mathbf{w}|s,\mathcal{D}_t}
    \,\mathrm{d}\mathbf{w} \nonumber\\
    &= \int \sum_{i=1}^n \hat{P}_i(s'|s,a) w_i \prob{\mathbf{w}|s,\mathcal{D}_t} \,\mathrm{d}\mathbf{w} = \sum_{i=1}^n \hat{P}_i(s'|s,a) \int w_i \prob{\mathbf{w}|s,\mathcal{D}_t} \,\mathrm{d}\mathbf{w} \nonumber\\
\label{eqn:mixture}
    &= \sum_{i=1}^n \hat{P}_i(s'|s,a)\, \EL{\mathbf{w}\sim \prob{\mathbf{w}|s,\mathcal{D}_t}}{w_i}.
\end{align}
In the following sections, we will instead refer to the following normalized form of (\ref{eqn:mixture})
\begin{equation}
\label{eqn:p_as_mixture}
    \prob{s'|s,a,\mathcal{D}_t} = \sum_{i=1}^n \hat{P}_i(s'|s,a) \,a_{t,i}(s),
\end{equation}
where $a_{t,i}(s) = \EL{\mathbf{w}\sim \prob{\mathbf{w}|s,\mathcal{D}_t}}{w_i} = \frac{\alpha_{t,i}(s)}{\sum_{j=1}^n \alpha_{t,j}(s)}$. Therefore, the posterior estimate of target dynamics (\ref{eqn:p_as_mixture}) can be interpreted as a contextual mixture of source task models.

In a tabular setting, it is feasible to maintain separate estimates of $a_{t,i}(s)$ per state using approximate Bayes' inference \citep{andrieu2003introduction}. However, maintaining such estimates for large or continuous state spaces presents inherent computational challenges. Fortunately, as (\ref{eqn:p_as_mixture}) showed, the posterior mean $\mathbf{a}_t(s)$ is a sufficient estimator of $\prob{s'|s,a,\mathcal{D}_t}$. Therefore, we can approximate $\mathbf{a}_t(s)$ directly using a feed-forward neural network $\mathbf{a}(s;\bm{\theta}_t)$ with parameters $\bm{\theta}_t$. The input of $\mathbf{a}(s;\bm{\theta})$ is a vectorized representation of $s$, and the outputs $z_{i}^a(s;{\bm{\theta}})$ are fed through the softmax function $a_{i}(s;\bm{\theta}) \propto {\exp{\left(z_{i}^a(s;{\bm{\theta}}\right))}}$ to guarantee that $\mathbf{a}_{t}(s)\geq \mathbf{0}$ and $\sum_{i=1}^n a_{t,i}(s) = 1$. Now it is no longer necessary to store all $\mathcal{D}_t$, since each sample can be processed online or in batches. Furthermore, since $\bm{\theta}_t$ is a neural network approximation of (\ref{eqn:p_as_mixture}), we can write $\prob{s' | s, a, \mathcal{D}_t} \simeq \prob{s' | s, a, \bm{\theta}_t}$ \citep{cybenko1989approximation}.

In order to learn the parameters $\bm{\theta}$, we can minimize the empirical negative log-likelihood function\footnote{Note that this equates to optimizing the posterior $\prob{\bm{\theta} | \mathcal{D}_t}$ with a uniform prior $\prob{\bm{\theta}}$.} using gradient descent, given by (\ref{eqn:p_as_mixture}) as:
\begin{align}
\label{eqn:loss}
    \mathcal{L}(\bm{\theta}) 
    &= -\probL{s'\sim P_{target}(s' | s,a)}{\mathcal{D}_t | \bm{\theta}} = -\sum_{(s,a,s')\in \mathcal{D}_t} \log\left(\sum_{i=1}^n \hat{P}_i(s'|s,a)\, a_{i}(s; \bm{\theta})\right).
\end{align}

The gradient of $\mathcal{L}(\bm{\theta})$ has a Bayesian interpretation. For one sample $(s,a,s')$, it can be written as
\begin{equation}
\label{eqn:theta_ascent}
    \nabla_{\bm{\theta}} \mathcal{L}(\bm{\theta}) = \sum_{i=1}^n \left(a_i(s; \bm{\theta})- p_i(s; \bm{\theta}) \right) \nabla_{\bm{\theta}} z_i^a(s;\bm{\theta}), 
    \quad p_i(s; \bm{\theta}) = \frac{\hat{P}_i(s'|s,a)\, a_i(s; \bm{\theta})}{\sum_{j=1}^n \hat{P}_j(s'|s,a) \,a_j(s; \bm{\theta})}.
\end{equation}
Here, we can interpret $a_i(s; \bm{\theta})$ as a prior. Once a new sample $(s,a,s')$ is observed, we compute the posterior $p_i(s; \bm{\theta})$ using Bayes rule, and $\bm{\theta}$ is updated according to the difference between prior and posterior, scaled by state features $z_i^a$. Hence, gradient updates in $\bm{\theta}$ space can be viewed as projections of posterior updates in $\mathbf{w}$ space. Regularization of $a_i(s; \bm{\theta})$ can also be easily incorporated by using informative priors $\prob{\bm{\theta}}$ (e.g. isotropic Gaussian, Laplace) in (\ref{eqn:loss}), and can lead to smoother posteriors.

Our approach is based on deep neural networks, so theoretical convergence is impossible to establish in general. However, we can still show that incorporating dynamics as part of the transfer learning process contextually can lead to better estimates of the value function, and hence better policies.
\begin{theorem}
\label{theorem:main}
    Consider an MDP $\langle \mathcal{S}, \mathcal{A}, P, R, \gamma\rangle$ with finite $\mathcal{S}$ and $\mathcal{A}$ and bounded reward $R : \mathcal{S} \to \mathbb{R}$. Let $\mathbf{R}$ be the reward function in vector form, $\hat{\mathbf{P}}^{\pi}$ be an estimate of the transition probabilities induced by a policy $\pi : \mathcal{S} \to \mathcal{A}$ in matrix form, and $\hat{\mathbf{V}}^{\pi}$ be the corresponding value function in vector form. Also, let $\mathbf{P}^{\pi}$ and $\mathbf{V}^\pi$ be the corresponding values under the true dynamics. Then for any policy $\pi$,
    \begin{equation*}
        \| \hat{\mathbf{V}}^{\pi} - \mathbf{V}^{\pi} \|_\infty \leq \frac{\gamma}{(1 - \gamma)^2} \,\| \mathbf{R} \|_\infty \,\| \hat{\mathbf{P}}^{\pi} - \mathbf{P}^{\pi} \|_\infty.
    \end{equation*}
\end{theorem}
A proof is provided in the Appendix. This result justifies our methodology of using source task dynamics similarity to guide state-dependent policy reuse, since actively minimizing the error term $\| \hat{\mathbf{P}}^{\pi} - \mathbf{P}^{\pi} \|_\infty$ can lead to significant improvements in transfer.

\subsection{Conditional RBF Network}
\label{subsec:rbf}

In continuous-state tasks with deterministic transitions, $P_i(s'|s,a)$ correspond to Dirac measures, in which case we only have access to models $\hat{f}_{i}(s,a)$ that predict the next state. In order to tractably update the mixture model in this setting, we assume that, given source task $i$ is the correct model of target dynamics, the probability of observing a transition from state $s$ to state $s'$ is a decreasing function of the prediction error $\|s'-\hat{s}'\|=\|s'-\hat{f}_{i}(s,a)\|$. More formally, given an arbitrarily small region $S=[s', s' + \mathrm{d}s']$,
\begin{equation}
\label{eqn:rbf}
    \prob{s'\in S |s,a,\mathbf{w},i} = \rho_i\left(\|s' - \hat{f}_{i}(s,a) \|\right) \,\mathrm{d}s',
\end{equation}
where $\rho_i : \mathbb{R} \to \mathbb{R}$ can be interpreted as a normalized\footnote{Technically, we could only require that $\rho_i \geq 0$, as the likelihood need not be a valid probability density.} \emph{radial basis function}. A popular choice of $\rho_i$, implemented in this paper, is the \emph{Gaussian kernel}, which for $\nu_i > 0$ is $\rho_i(r) \propto \exp{\left(-\nu_i r^2\right)}$. In principle, $\nu_i$ could be modeled as an additional output of the mixture model, $\nu_i(\bm{\theta})$, and learned from data \citep{bishop1994mixture}, although we treat it as a constant in this paper for clarity.

By using (\ref{eqn:rbf}) and following the derivations leading to (\ref{eqn:mixture}), we obtain the following result in direct analogy to (\ref{eqn:p_as_mixture})
\begin{equation}
\label{eqn:p_as_mixture_kernel}
    \prob{s'\in S|s,a,\mathcal{D}_t} = \sum_{i=1}^n \rho_i\left(\|s' - \hat{f}_{i}(s,a) \|\right) a_{t,i}(s) \,\mathrm{d}s'.
\end{equation}
Consequently, the results derived in the previous sections, including the mixture model and loss function for $\mathbf{a}(s; \bm{\theta}_t)$, hold by replacing $\hat{P}_i(s'|s,a)$ with $\rho_i\left(\|s' - \hat{f}_{i}(s,a) \|\right)$. Furthermore, since (\ref{eqn:p_as_mixture_kernel}) approximates the target dynamics as a mixture of kernel functions, it can be viewed as a conditional analogue of the \emph{radial basis function network} \citep{broomhead1988radial}. It remains to show how to make use of this model and the source policy library to solve a new target task.

\subsection{Policy Transfer} 
\label{subsec:mars}

\paragraph{Policy Shaping} 
The most straightforward approach for policy reuse is to reshape the policy directly \citep{griffith2013policy}. In other words, the agent samples a source policy at random at each time step $t$ according to $\mathbf{a}(s_t; \bm{\theta}_t)$, and follows it in state $s_t$. To allow for exploration, the agent only takes advice with probability $p_t \in [0, 1]$, otherwise following a target exploration policy \cite{fernandez2006probabilistic,li2018optimal}. This resulting behaviour policy is suitable for any off-policy RL algorithm. We call this \emph{\textbf{M}odel-\textbf{A}ware \textbf{P}olicy \textbf{S}haping for \textbf{E}xploration} (MAPSE), and include the pseudocode of this procedure in the Appendix.

However, such an approach has several shortcomings. Firstly, it is not clear how to anneal $p_t$, since $\mathbf{a}(s; \bm{\theta}_t)$ is learned over time and is non-stationary. Secondly, using the recommended actions too often can lead to poor test performance, since the agent may not observe sub-optimal actions enough times to learn to avoid them in testing. Finally, since efficient credit assignment is particularly difficult in sparse reward problems \citep{seo2019}, it may limit the effectiveness of action recommendation. 

\paragraph{Reward Shaping} 
An alternative approach for incorporating policy advice is to reshape the original reward function \citep{brys2015policy}. More specifically, the original reward signal $R_t(s,a,s')$ at time $t$ is modified to a new signal $R_t'(s,a,s') = R_t(s,a,s') + c F_t(s,a,s',a')$, where
\begin{equation}
\label{eqn:rs}
    F_t(s,a,s',a') = \gamma \Phi_{t+1}(s',a') - \Phi_t(s,a);
\end{equation}
here, $c > 0$ defines the strength of the shaped reward signal, and can be tuned for each problem or empirically scaled to the magnitude of the original reward function. The potential $\Phi_t(s,a)$ is chosen to be the posterior probability $\prob{a | s, \mathcal{D}_t}$ that action $a$ would be recommended by a source policy in state $s$ at time $t$. By repeating the derivations leading to (\ref{eqn:mixture}), we can derive an expression for $\Phi_t(s,a)$:
\begin{align}
\label{eqn:potential}
    \Phi_t(s,a) = \prob{a | s, \mathcal{D}_t} &= \int \sum_{i=1}^n \prob{a | s, \mathcal{D}_t, \mathbf{w}, i} \prob{i | \mathbf{w}, s} \prob{\mathbf{w} | s, \mathcal{D}_t} \,\mathrm{d}\mathbf{w} \nonumber \\ 
    &= \sum_{i=1}^n \prob{a = \pi_i^*(s)}\, a_i(s; \bm{\theta}_t).
\end{align}

Note that (\ref{eqn:potential}) reduces to \citet{brys2015policy} when $n=1$ and source policies are deterministic. Unlike MAPSE, this approach can also be applied on-policy, and preserves policy optimality \citep{devlin2012dynamic}. It can also be applied for continuous control by modelling $\prob{a = \pi_i^*(s)}$ using kernels as in Section~\ref{subsec:rbf}. We call this approach \emph{\textbf{M}odel-\textbf{A}ware \textbf{R}eward \textbf{S}haping} (MARS). A complete description of the training procedure of MARS and MAPSE is provided in the Appendix.

\paragraph{Remarks} 
The proposed framework is general and modular and can be combined with most RL algorithms and MBRL. Furthermore, the computational cost of processing each sample is a linear function of the cost of evaluating the source task dynamics and policies, which can be efficiently implemented using neural networks. In a practical implementation, $\bm{\theta}$ would be trained with a higher learning rate or larger batch size than the target policy, to make use of source task information earlier.

Many extensions to the current framework are possible. For instance, to reduce the effect of negative transfer, it is possible to estimate the target task dynamics $\hat{P}_{target}(s'|s,a)$ or $\hat{f}_{target}(s,a)$, and include it as an additional ($n + 1$)-st component in the mixture (\ref{eqn:p_as_mixture}). If this model can be estimated accurately, it can also be used to update the agent directly as suggested in Section~\ref{sec:prelim}. Further improvements for MARS could be obtained by learning a secondary Q-value function for the potentials \citep{harutyunyan2015expressing}. We do not investigate these extensions in this paper, which can form interesting topics for future study.

\section{Empirical Evaluation}
\label{sec:empirical}

In this section, we evaluate, empirically, the performance of both MAPSE and MARS in a typical RL setting. In particular, we aim to answer the following questions:
\begin{enumerate}
    \item Does $\mathbf{a}(s; \bm{\theta}_t)$ learn to select the most relevant source task(s) in each state?
    \item Does MARS (and possibly MAPSE) achieve better test performance, relative to the number of transitions observed, over existing baselines?
    \item Does MARS lead to better transfer than MAPSE?
\end{enumerate}

In order to answer these questions, we consider tabular \textbf{Q-learning} \citep{watkins1992q} and \textbf{DQN} \citep{mnih2015human} with MAPSE and MARS. To ensure fair comparison with relevant baselines, we include one state-of-the-art context-free and contextual policy reuse algorithm:

\begin{enumerate}
    \item \textbf{CAPS:} a contextual option-based algorithm recently proposed in \citet{li2019context};
    \item \textbf{UCB:} a context-free UCB algorithm proposed in \citet{li2018optimal};
    \item \textbf{$\Phi$1, $\Phi$2\dots:} PBRS derived from each source policy $i$ as suggested in \citet{brys2015policy};
    \item \textbf{Q:} Q-learning or DQN with no transfer.
\end{enumerate}

To help us answer the research questions above, we consider three variants of existing problems, \textbf{Transfer-Maze}, \textbf{Transfer-CartPole}, and \textbf{Transfer-SparseLunarLander}, that are explained in the subsequent subsections. Further implementation details are provided in the Appendix.

\subsection{Transfer-Maze}

The first experiment consists of a 30-by-30 discrete maze environment with four sub-rooms. The four possible actions \texttt{$\lbrace$left, up, right, down$\rbrace$} move the agent one cell in the corresponding direction, but have no effect if the destination contains a wall. The agent incurs a penalty of -0.02 for hitting a wall, and otherwise -0.01. The goal is to get to a fixed destination cell in the least number of steps, at which point the agent receives a reward of +1.0. The source tasks each correctly model the interior of one room, so that only a context-aware algorithm can learn to utilize the source task knowledge correctly. A full description of this task is provided in the Appendix.

Here, we use Q-learning to learn optimal source and target policies. The dynamics $\hat{f}_i(s,a)$ are trained using lookup tables and hence $\hat{P}_i(s'|s,a)=\mathbbm{1}\lbrace s' = \hat{f}_i(s,a)\rbrace$. The experiment is repeated 20 times and the aggregated results are reported in Figure~\ref{fig:maze_learning}. Note that we have omitted the plot for $\Phi1,\Phi2\dots$ since convergence could not be obtained after $200,000$ steps using only single policies. Figure~\ref{fig:mars_maze_heatmap} plots the state-dependent posterior learned over time on a single trial. 

\subsection{Transfer-CartPole}

We next consider a variation of the continuous-state CartPole control problem from the OpenAi-Gym benchmark suite \citep{brockman2016openai}, where the force $F(x)$ applied to the cart is not constant, but varies with cart position $x \in [-2.4, +2.4]$ according to the equation $F(x) = 35 \sqrt{\frac{1 + 36}{1 + 36 \cos^2(5 x)}} \,\cos(5 x) + 40$. One way to interpret this is that the surface is not friction-less, but contains slippery ($F(x) \approx 75$) and rough ($F(x) \approx 5$) patches. To learn better policies, the agent can apply half or full force to the cart in either direction (4 possible actions). As a result, the optimal policy in each state depends on the surface. The problem is made more difficult by uniformly initializing $x \in [-1.5, +1.5]$, to require the agent to generalize control to both surfaces. In the first two source tasks, agents balance the pole only on rough ($F(x)=5$) and slippery ($F(x)=75$) surfaces, respectively. In the third source task, the pole length is doubled and $F(x)=20$.

\begin{figure}[!htb]
    \centering
        \begin{subfigure}[tb]{0.1925\textwidth}
        \centering
            \includegraphics[width=0.95\linewidth]{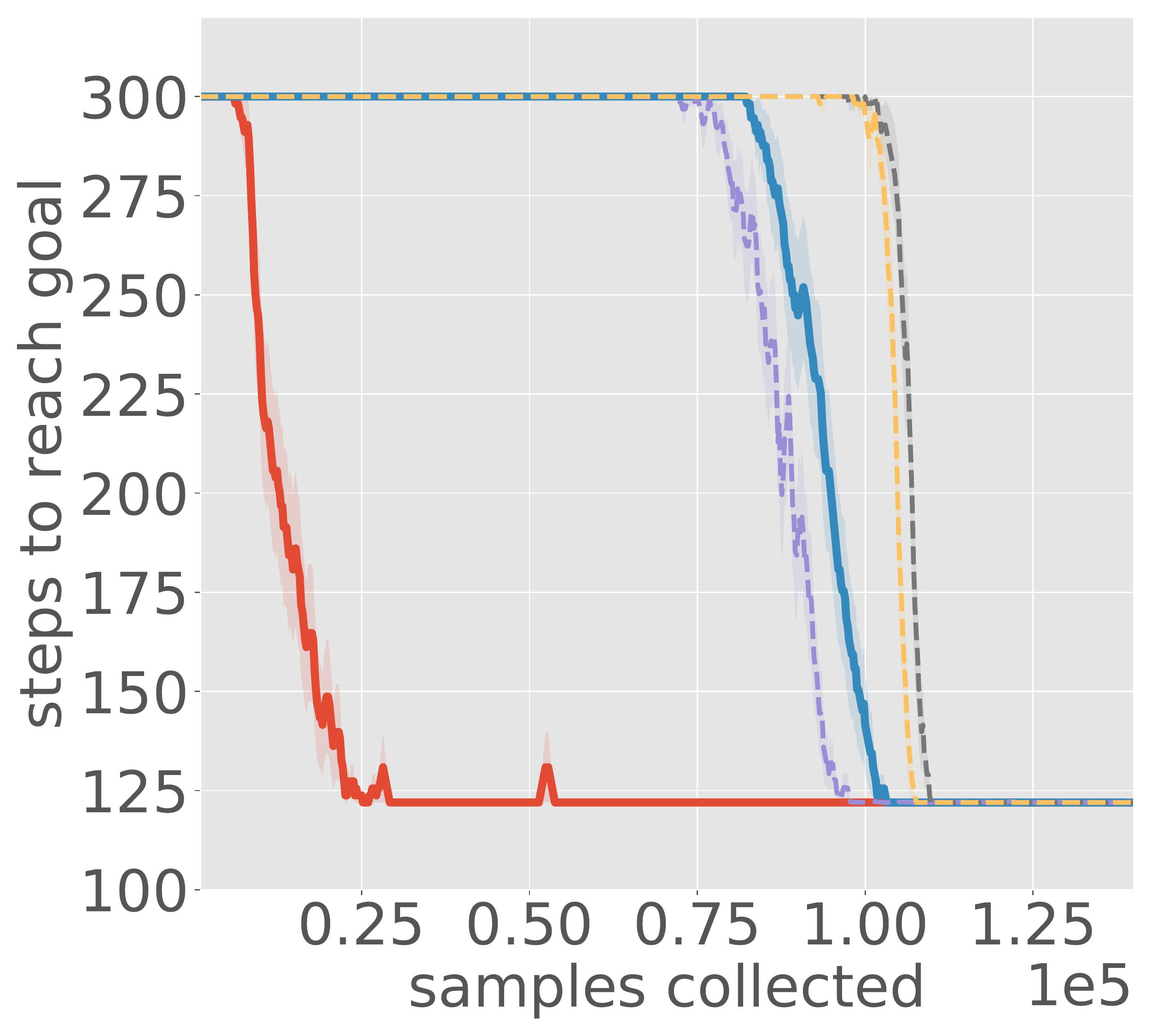}
            \caption{Transfer-Maze}
            \label{fig:maze_learning}
        \end{subfigure}\hfill
        \begin{subfigure}[tb]{0.395\textwidth}
        \centering
            \includegraphics[width=0.475\linewidth]{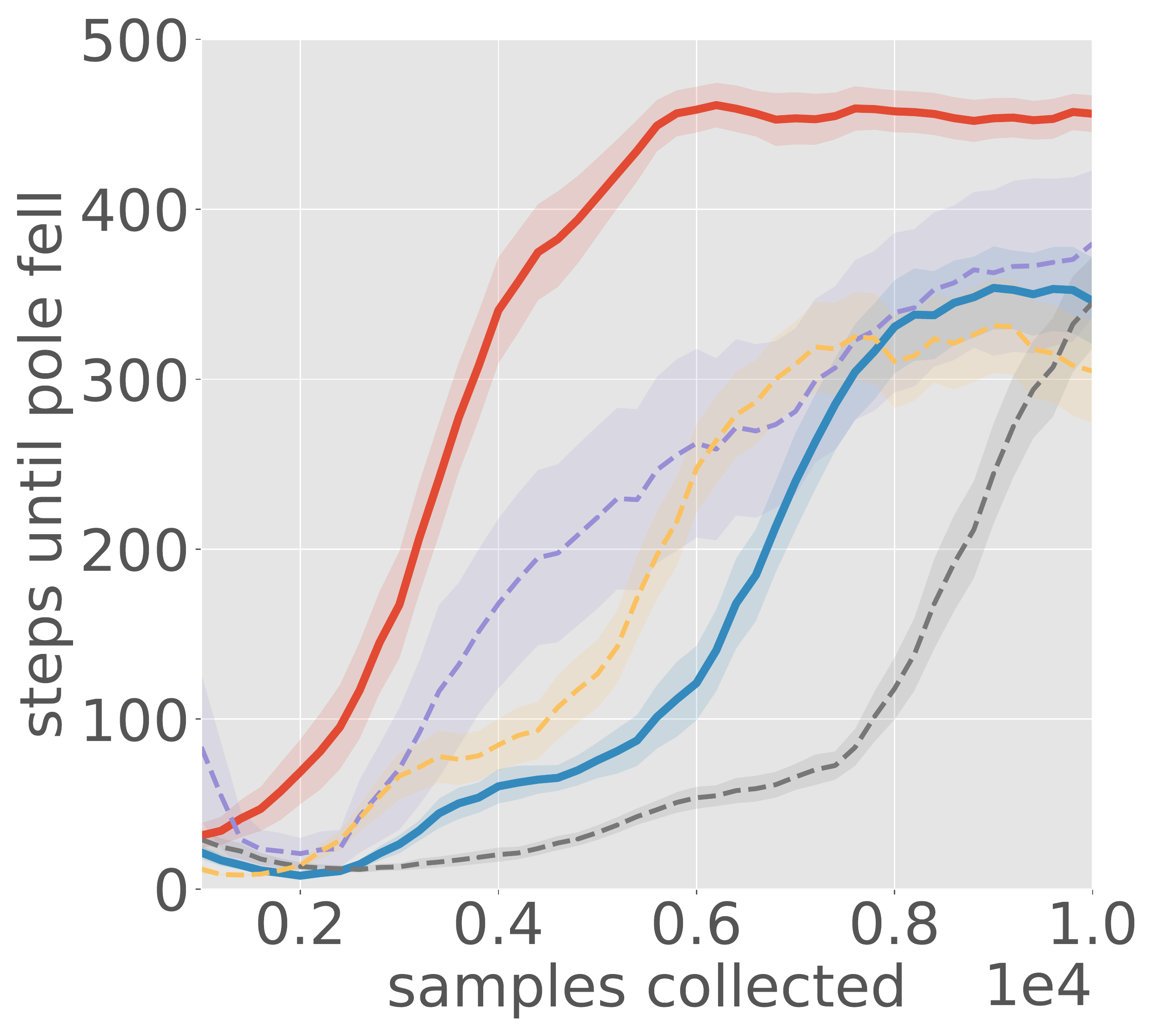}
            \includegraphics[width=0.475\linewidth]{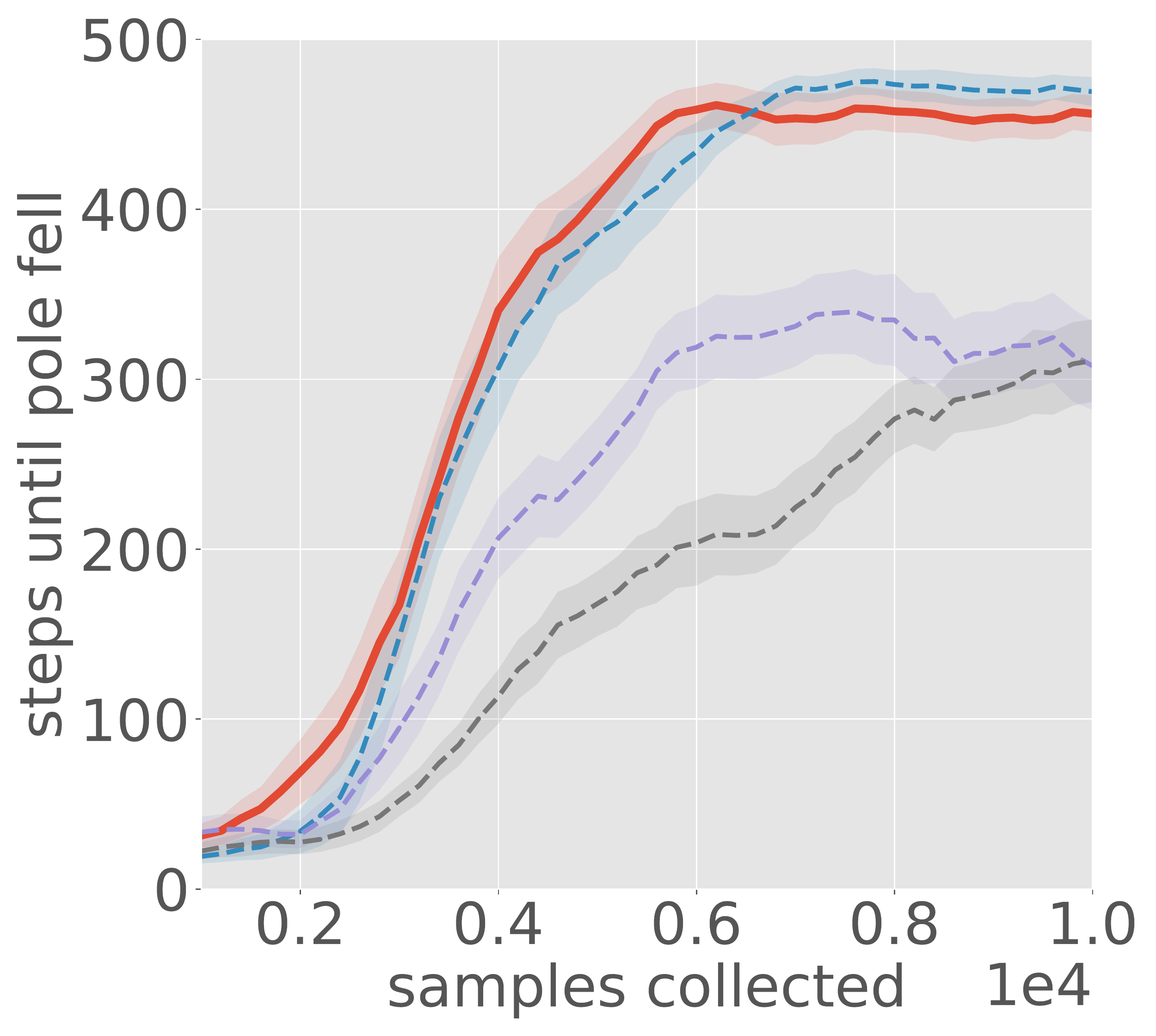}
            \caption{Transfer-CartPole}
            \label{fig:cartpole_learning}
        \end{subfigure}\hfill
        \begin{subfigure}[tb]{0.395\textwidth}
            \centering
            \includegraphics[width=0.475\linewidth]{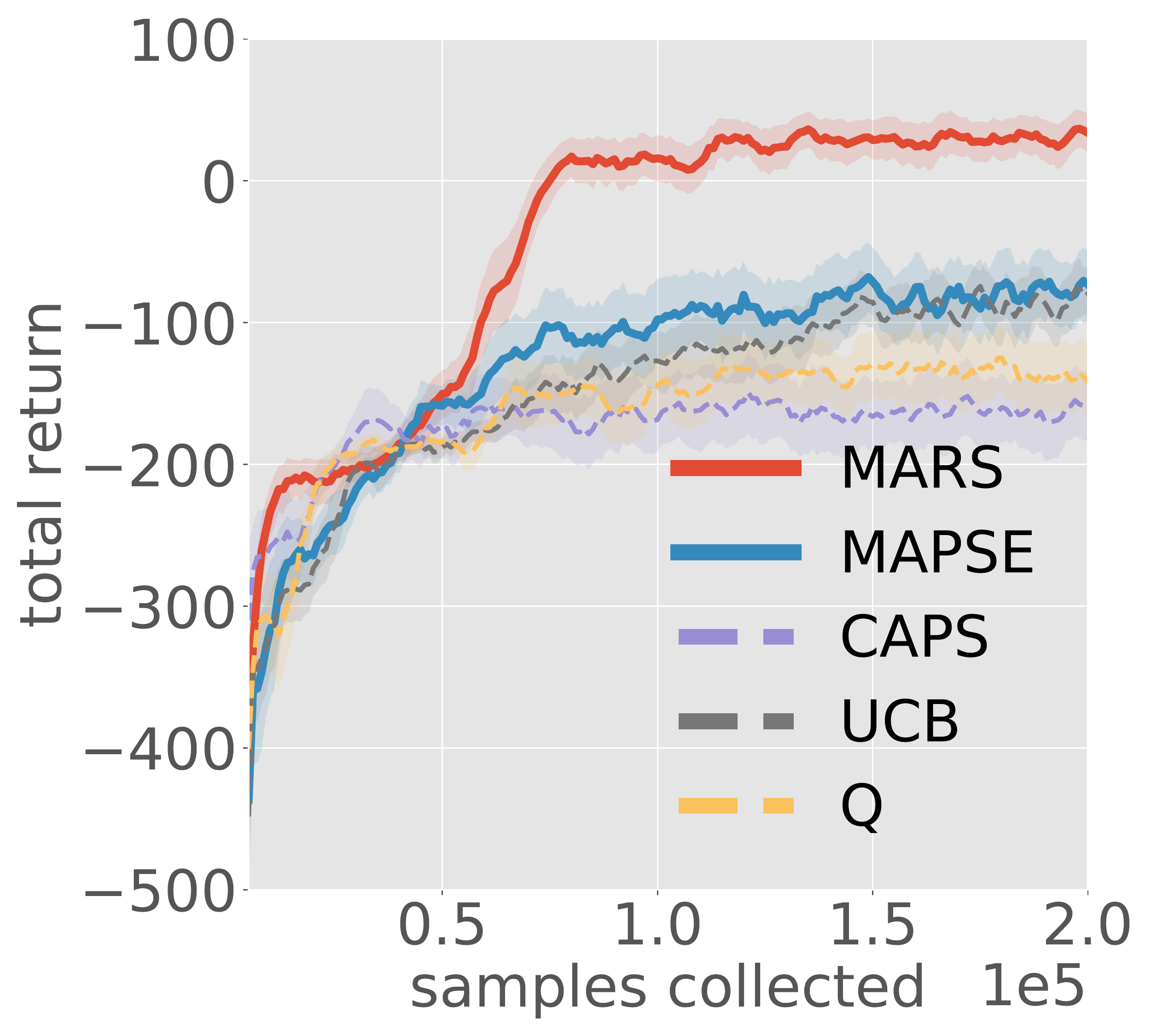}
            \includegraphics[width=0.475\linewidth]{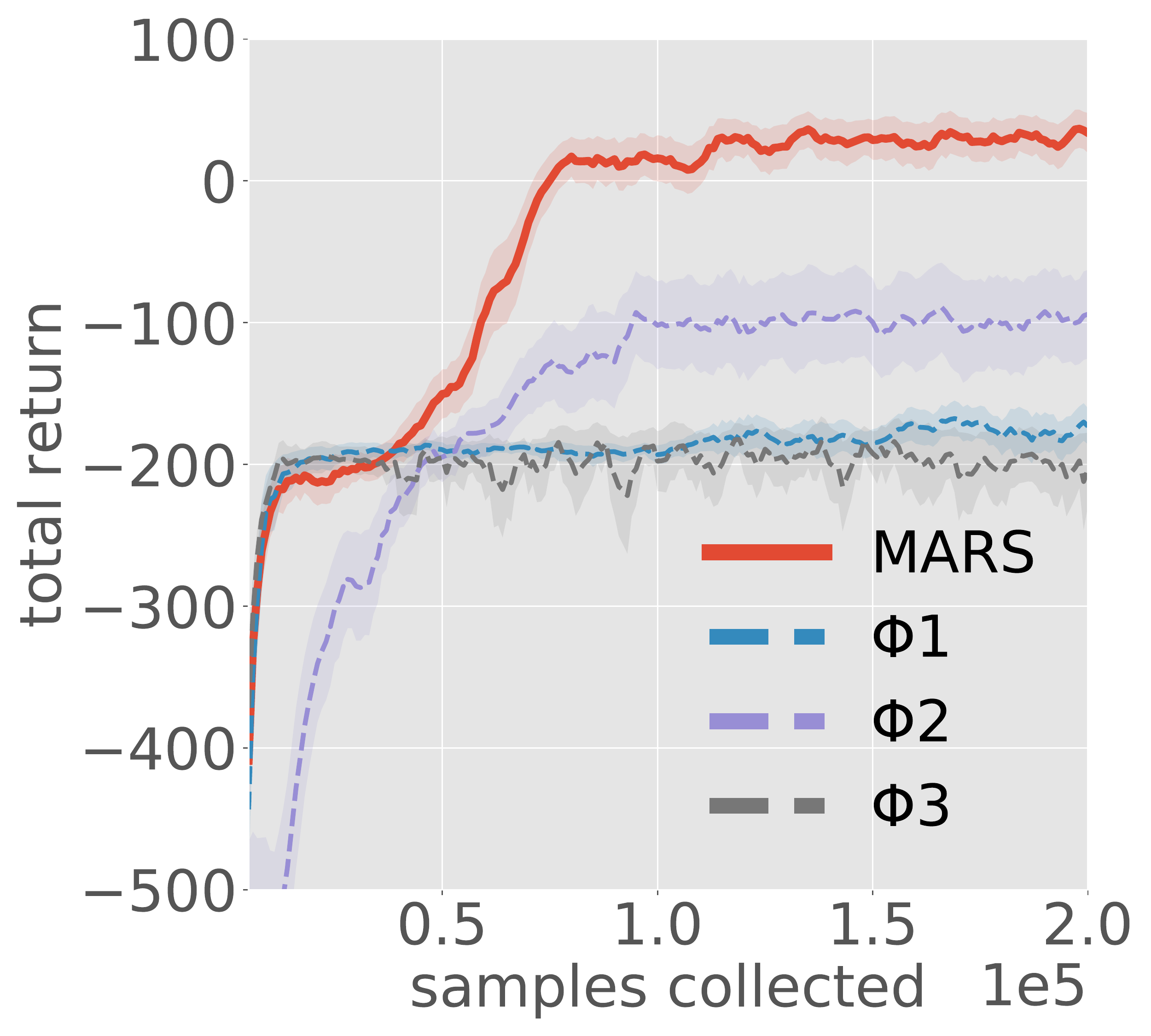}
            \caption{Tr.-SparseLunarLander}
            \label{fig:lander_learning}
        \end{subfigure}
        \caption{Smoothed mean test performance using the greedy policy: (a) number of steps to reach goal (b) number of steps balanced (c) total return and standard error over number of training samples, over 20 trials for Transfer-Maze and Transfer-CartPole and 10 trials for Transfer-SparseLunarLander.}
        \label{fig:learning_curve}
\end{figure}

Following \citet{mnih2015human}, Q-values are approximated using feed-forward neural networks, and we use randomized experience replay and target Q-network with hard updates. State dynamics $\hat{f}_i(s,a)$ are parameterized as feed-forward neural networks $f_{\bm{\phi}_i}(s,a)$ and trained using the MSE loss $\frac{1}{|\mathcal{B}|}\sum_{(s,a,s') \in \mathcal{B}}\|s' - f_{\bm{\phi}_i}(s,a) \|^2$ using batches $\mathcal{B}$ drawn randomly from the buffer. To learn $\mathbf{a}(s; \bm{\theta})$, the likelihood is estimated using Gaussian kernels with fixed $\nu$. For CAPS, we follow \citet{li2019context} and only train 
the last layer when learning the option termination function, whose learning rate is selected using validation from $\lbrace 0.01, 0.001, 0.0001\rbrace$. 
The test performance is illustrated in Figure~\ref{fig:cartpole_learning}. Figure~\ref{fig:mars_cartpole_heatmaps} plots the state-dependent posterior learned over time.

\subsection{Transfer-SparseLunarLander}

The final experiment consists of a variation of the {LunarLander-v2} domain from OpenAi-Gym, which involves landing a spacecraft safely on a lunar surface. Here, rewards are deferred until the end of each episode. This is a high-dimensional continuous stochastic problem with sparse reward, and representative of real-world problems where it is considerably difficult to learn accurate dynamics. The first source task teaches the lander to hover above the landing pad at a fixed region in space ($x \in [-0.1, +0.1], \,y \in [0.3, 0.5]$), and fails if the lander gets too close to the ground. The second source task places the lander at a random location ($x \in [-0.5, +0.5],\, y = 0.4$) above the landing pad, and the agent learns to land the craft safely. The third source task is equivalent to LunarLander-v2, except the craft's mass is only 10\% of the original mass. A successful transfer experiment, therefore, should learn to transfer skills from the hover and land source tasks depending on altitude, and try to avoid the risky policy for landing the lighter spacecraft.

\begin{figure}[!htb]
    \centering
    \begin{subfigure}{0.375\textwidth}
    \centering
    \begin{tabular}[b]{cc} 
         \includegraphics[width=0.475\linewidth]{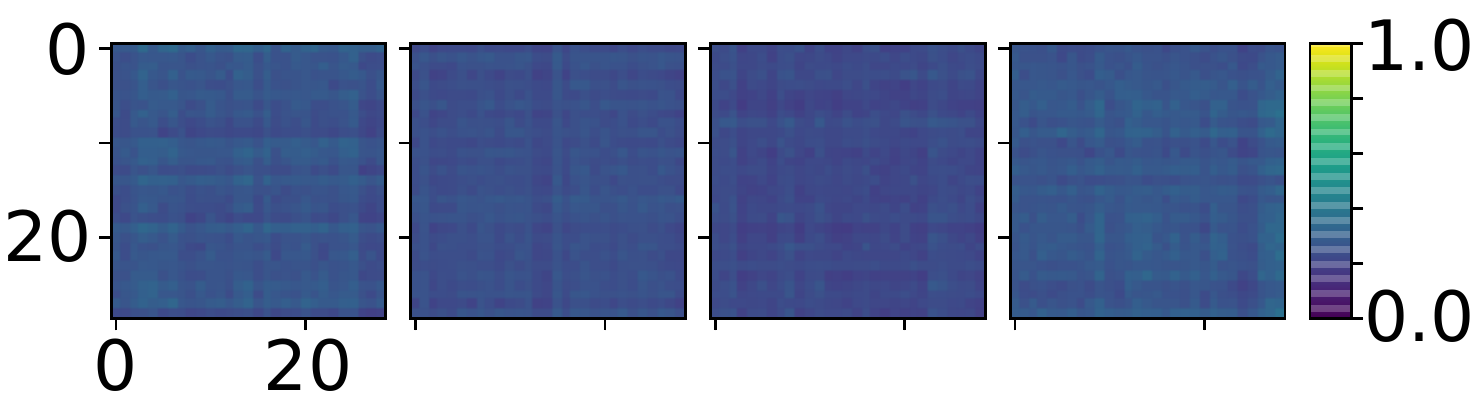} & \includegraphics[width=0.475\linewidth]{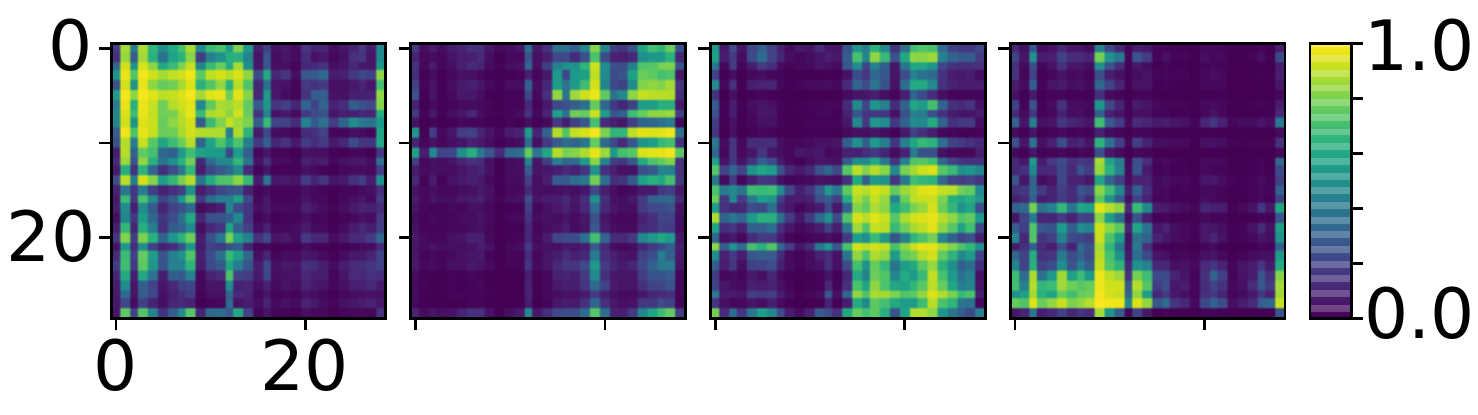} \\
         \includegraphics[width=0.475\linewidth]{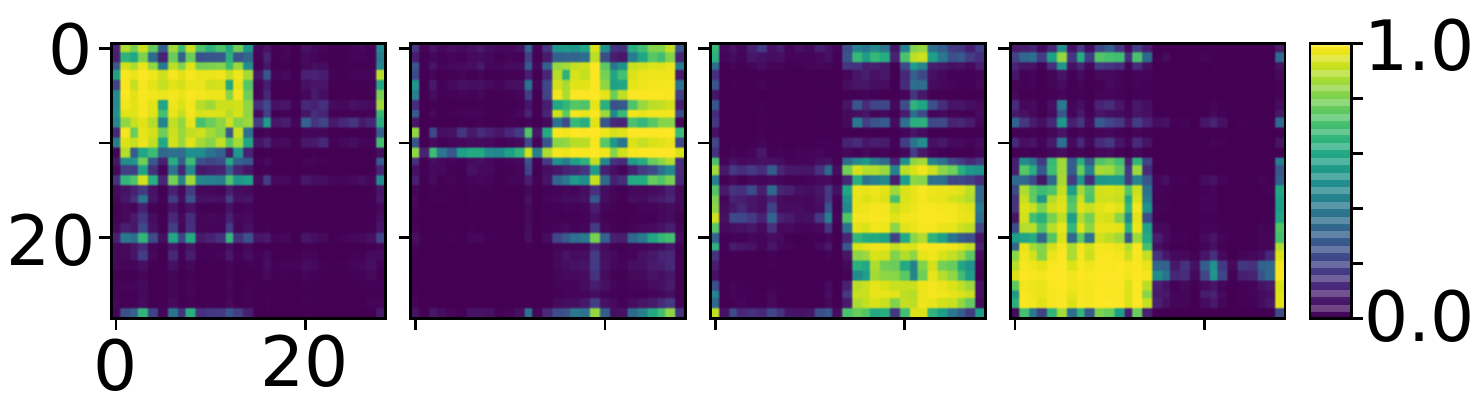} & \includegraphics[width=0.475\linewidth]{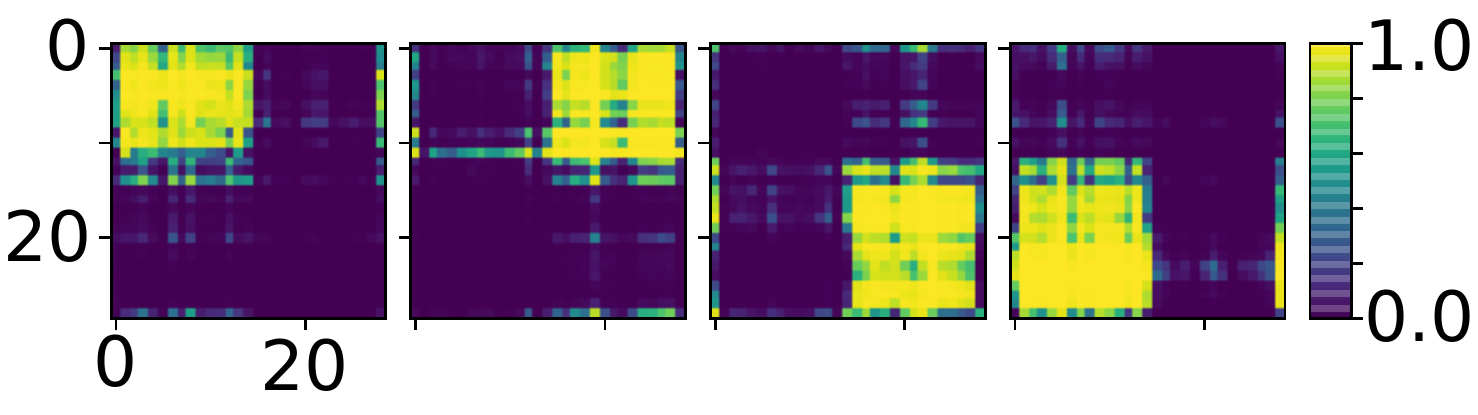} \\
         \includegraphics[width=0.475\linewidth]{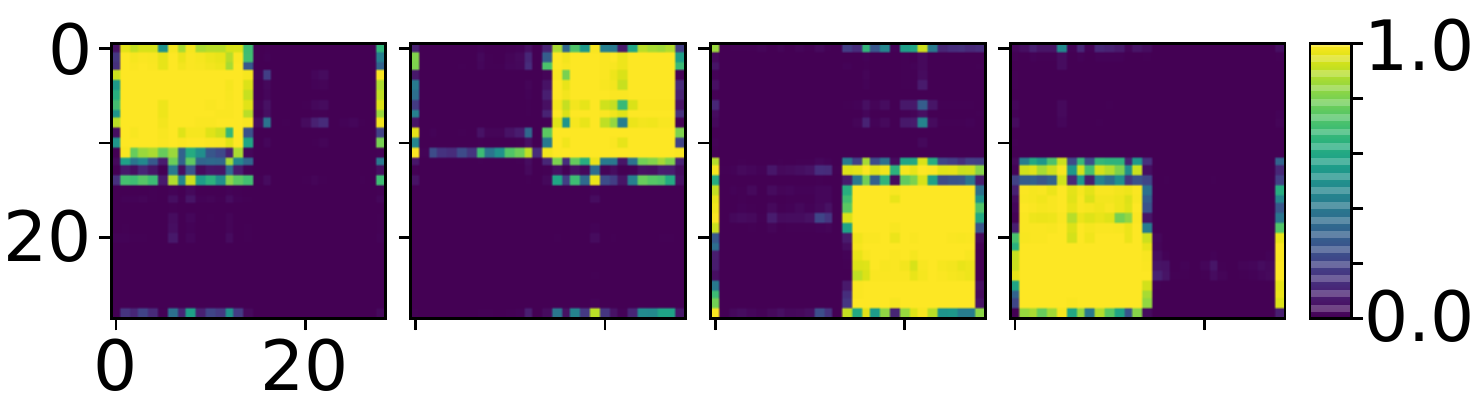} & \includegraphics[width=0.475\linewidth]{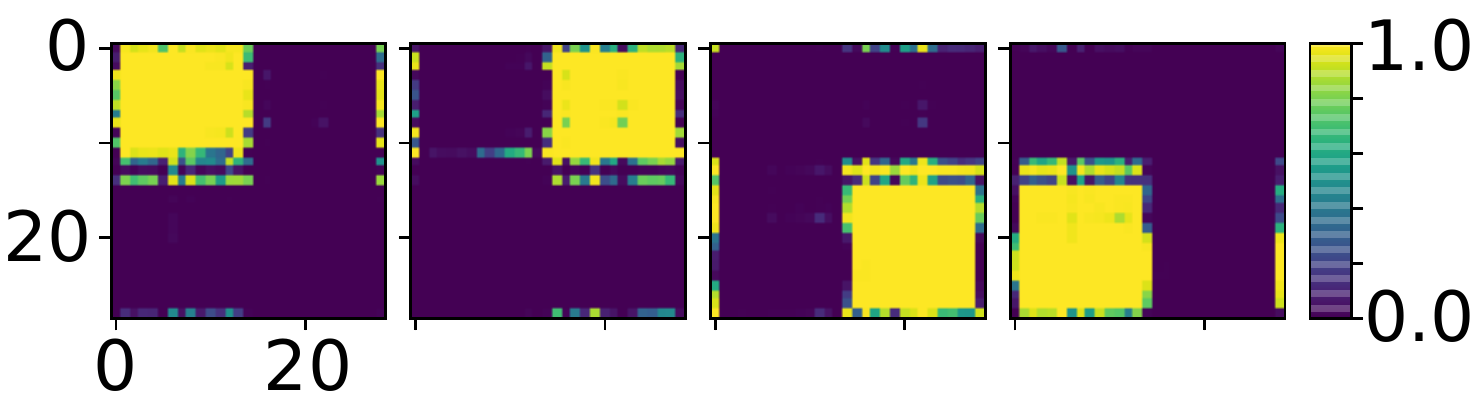}
        \end{tabular}
    \caption{Transfer-Maze}
    \label{fig:mars_maze_heatmap}
    \end{subfigure}\hfill%
    \begin{subfigure}{0.32\textwidth}
    \centering
    \begin{tabular}[b]{cc} 
        \includegraphics[width=0.475\linewidth]{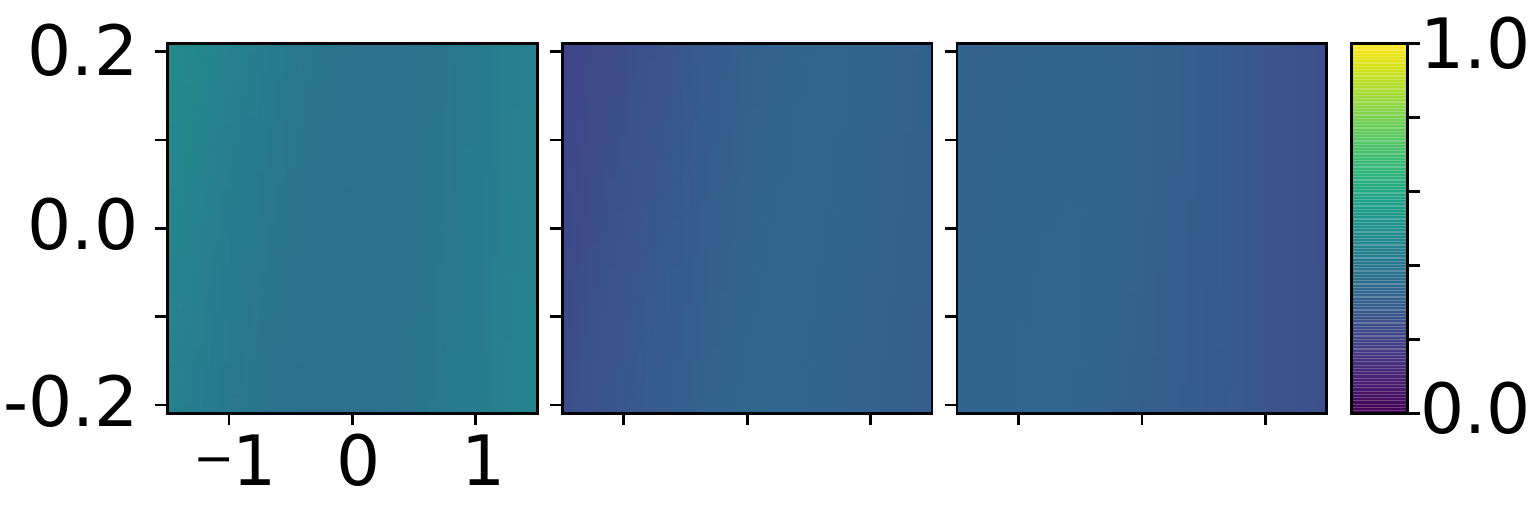} & \includegraphics[width=0.475\linewidth]{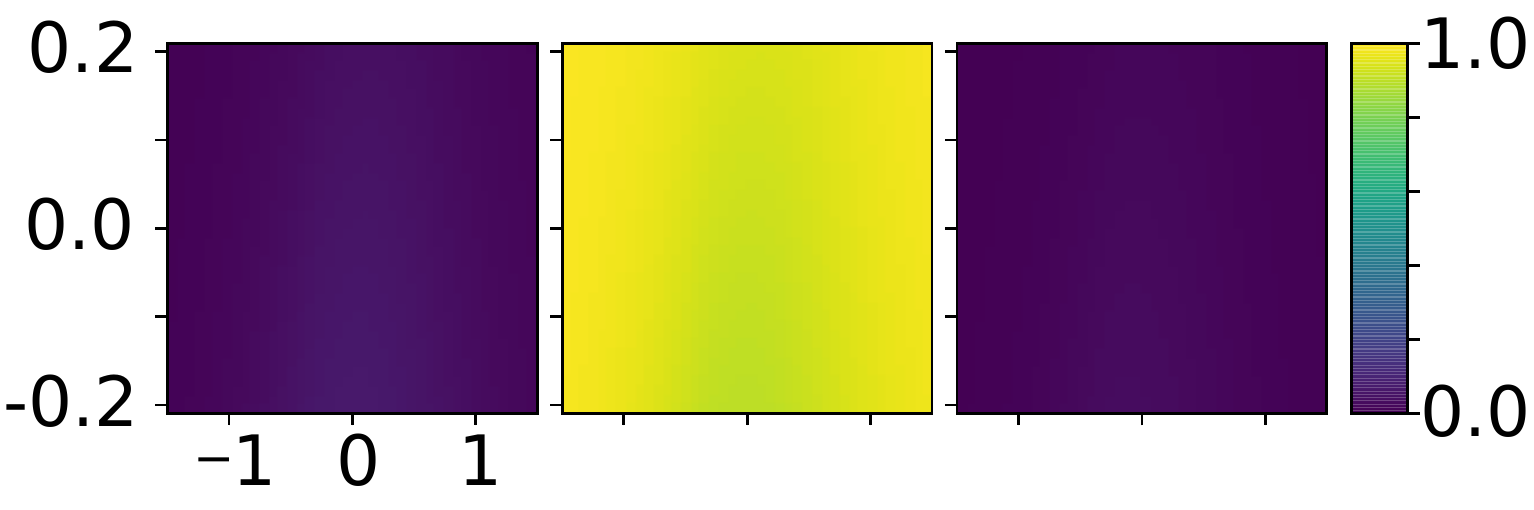} \\
        \includegraphics[width=0.475\linewidth]{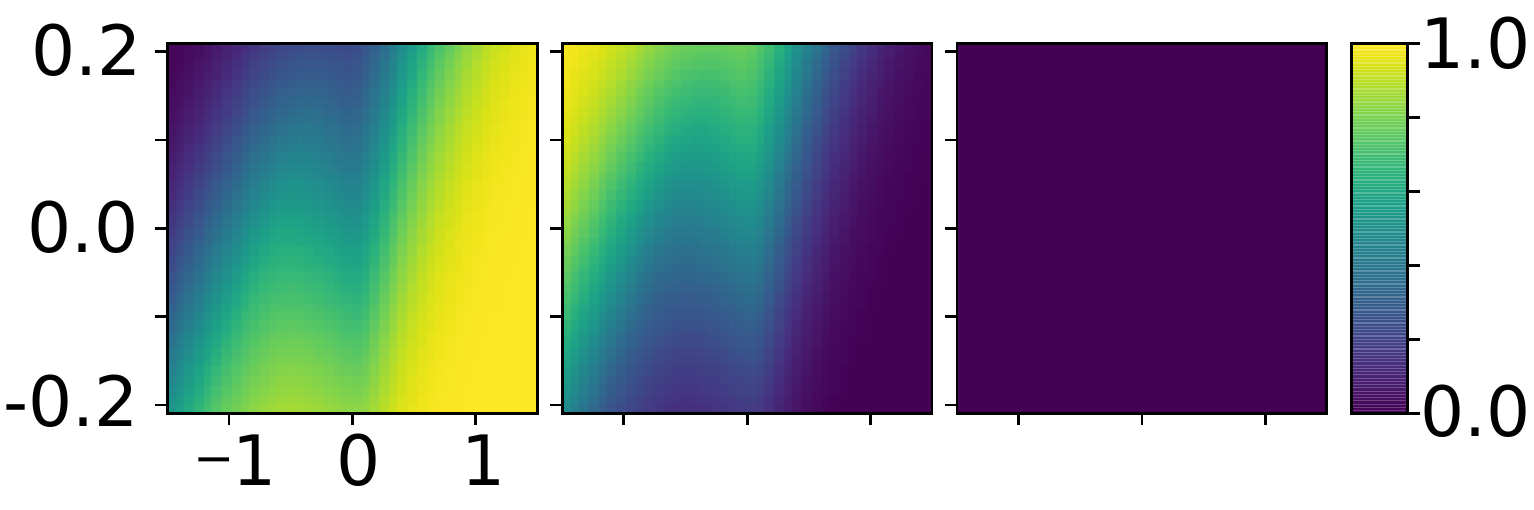} & \includegraphics[width=0.475\linewidth]{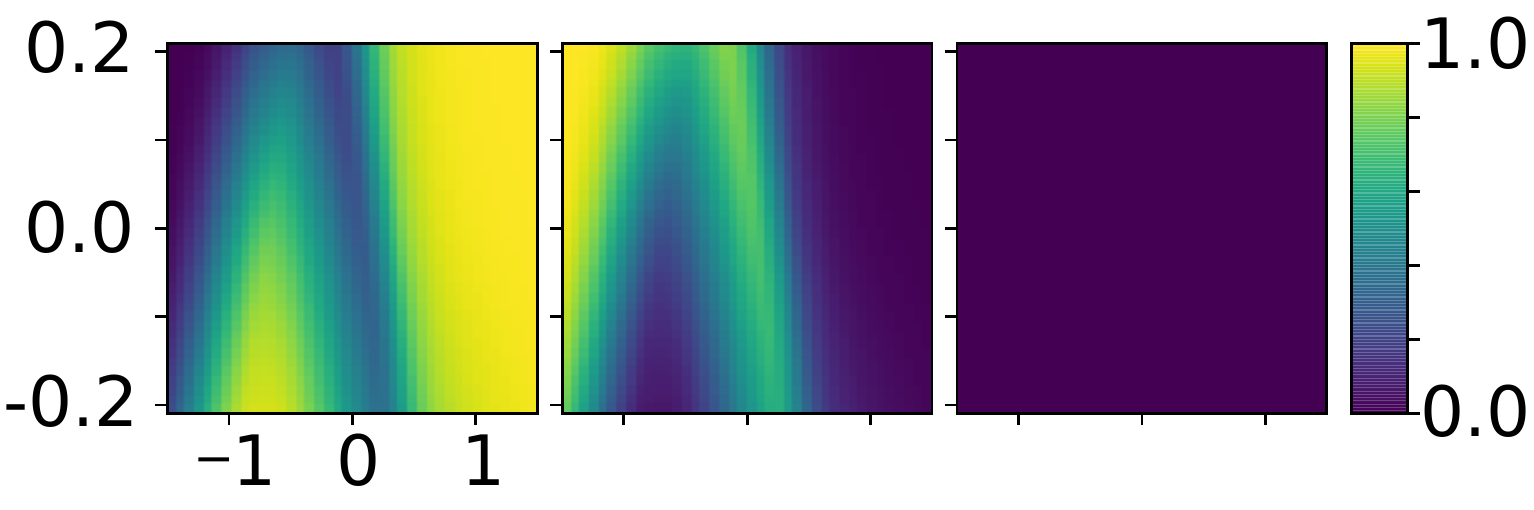} \\
        \includegraphics[width=0.475\linewidth]{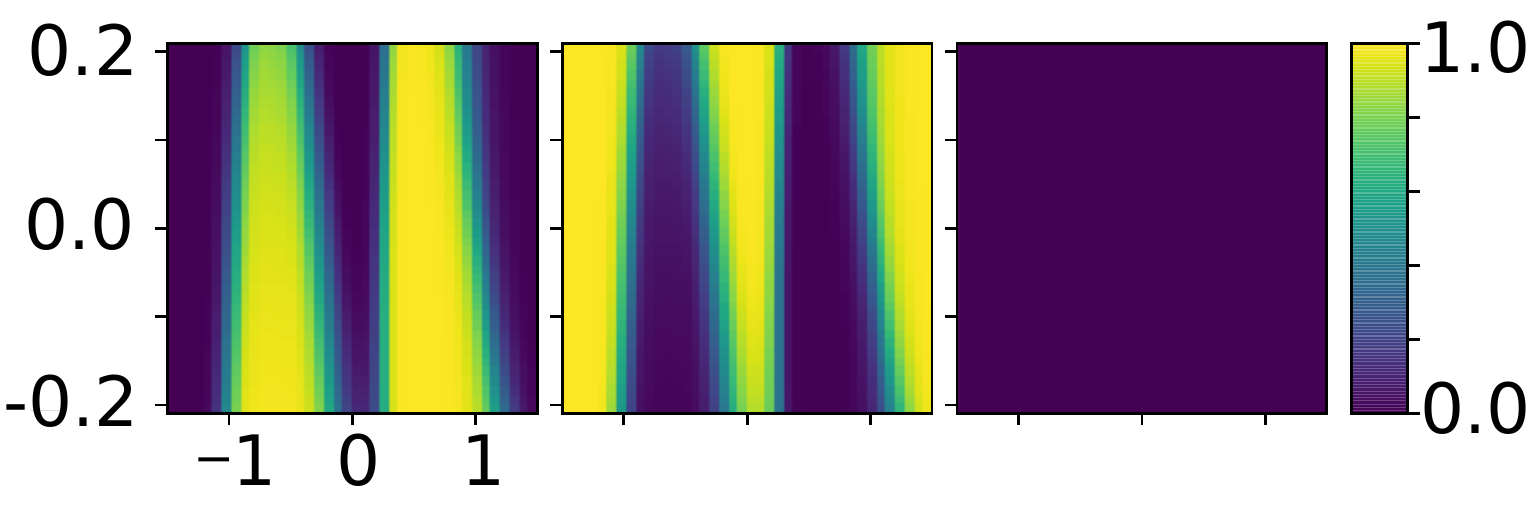} & \includegraphics[width=0.475\linewidth]{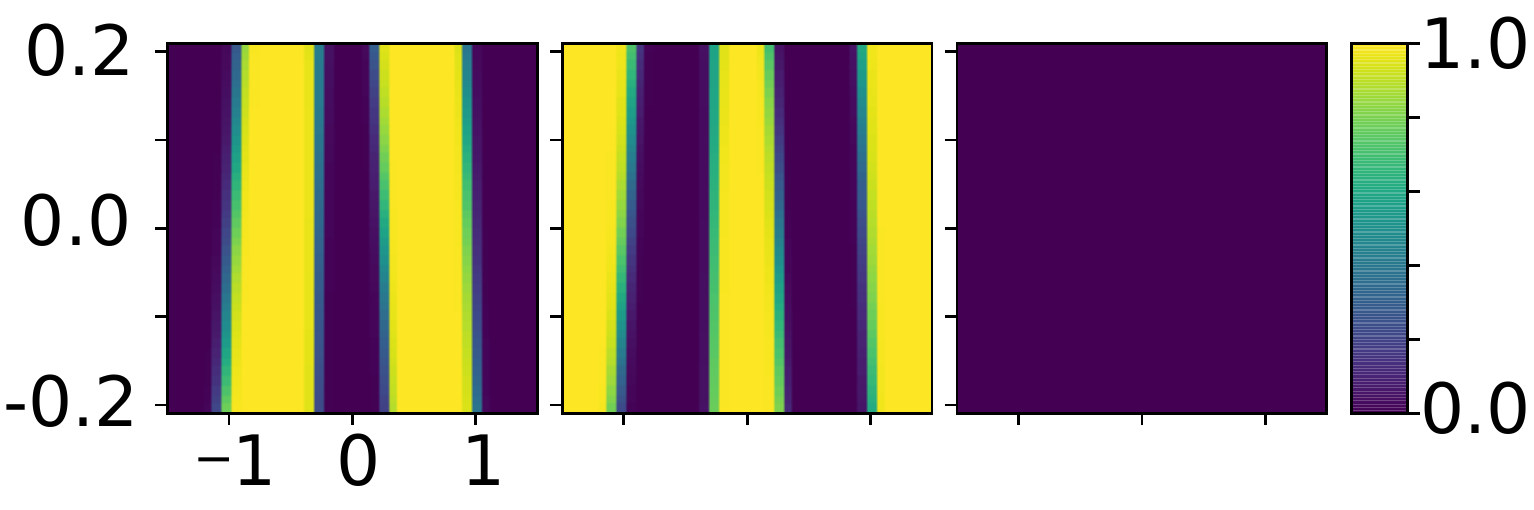}
    \end{tabular}
    \caption{Transfer-CartPole}
    \label{fig:mars_cartpole_heatmaps}
    \end{subfigure}\hfill%
    \begin{subfigure}{0.225\textwidth}
    \centering
        \begin{tabular}[b]{cc}
        \includegraphics[width=0.4\linewidth]{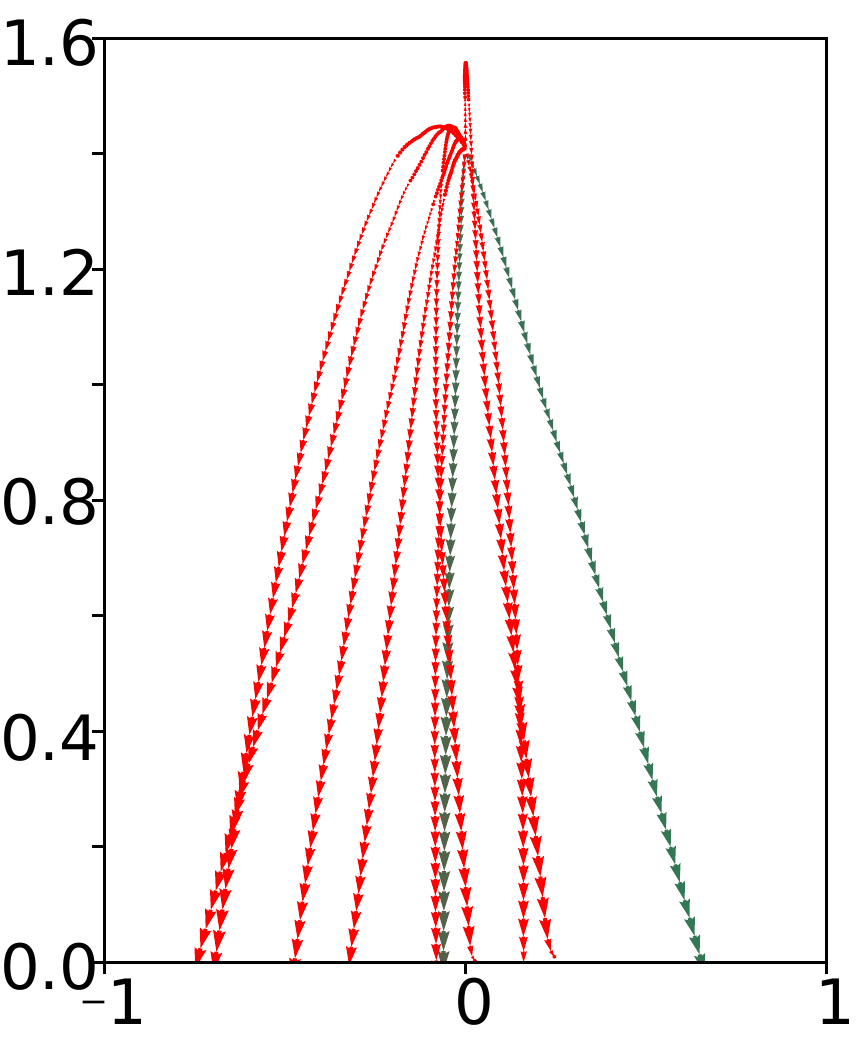} &
        \includegraphics[width=0.4\linewidth]{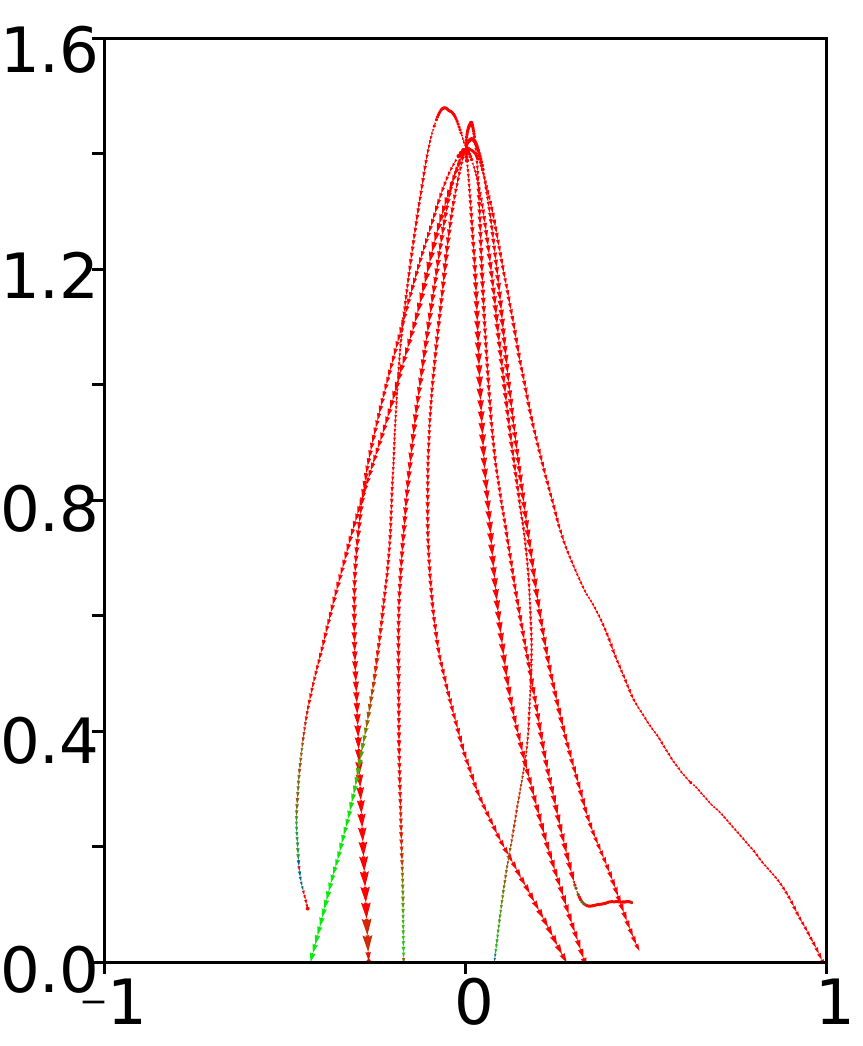} \\
        \includegraphics[width=0.4\linewidth]{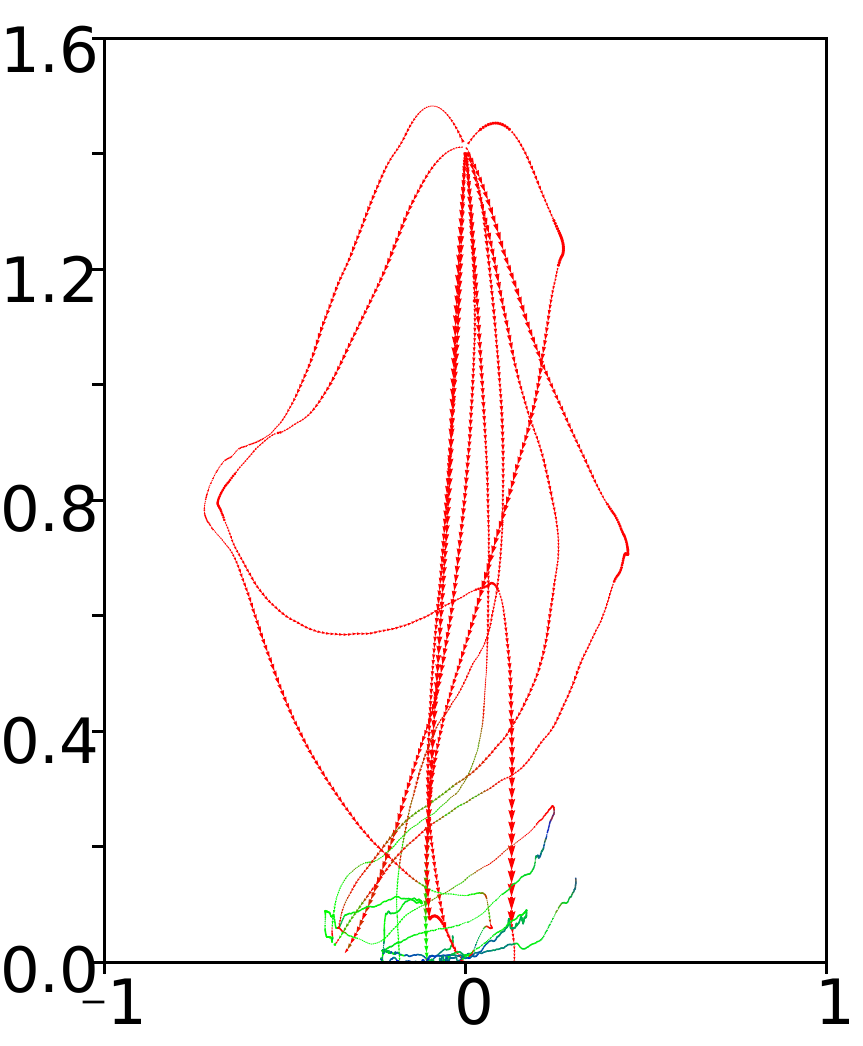}&
        \includegraphics[width=0.4\linewidth]{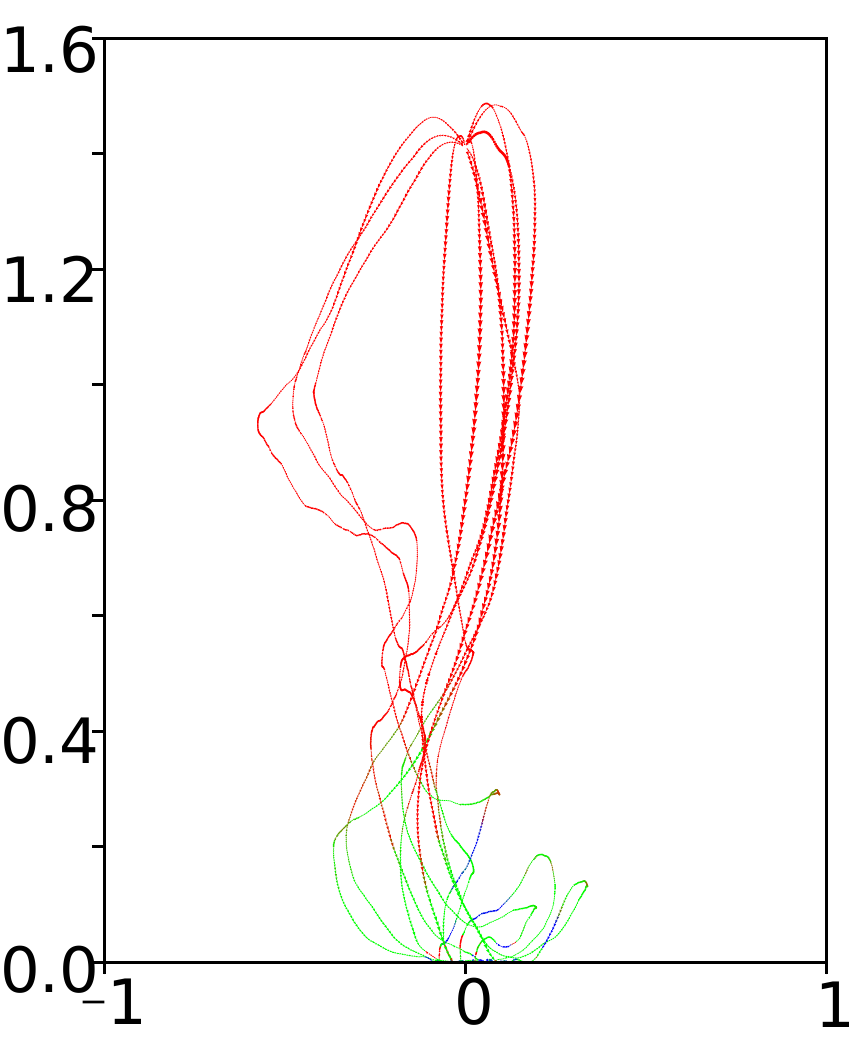}
        \end{tabular}
    \caption{Tr.-SparseLander}
    \label{fig:mars_lander_trajectories}
    \end{subfigure}
    \caption{Visualizing contextual transfer: (a) Probability assigned to each source task in each $(x,y)$ position of the Transfer-Maze domain in a single trial of MARS (results for MAPSE are similar) after collecting 0, 5K, 10K, 20K, 50K and 100K samples (left to right). (b) Probability assigned to each source task in each state of the Transfer-CartPole task in a single trial of MARS after 0, 100, 500, 1K, 2.5K and 5K training samples (left to right). x and y axes are cart position ($x$), averaged over $[-0.5, +0.5]$, and pole angle ($\theta$), respectively. (c) The lander position during 10 training episodes, collected after 0, 10K, 25K and 50K training samples. Colors indicate the source task mixture $\mathbf{a}(s; \bm{\theta}_t)$ where red corresponds to Hover, green to Land and blue to Low-Mass. It is clear that in each domain, MARS has learned to contextually transfer information from different source tasks in different states.}
    \label{fig:learning_representation}
\end{figure}

To solve this problem, we use the same setup as in Transfer-CartPole. Here, state transitions are stochastic and the moon surface is generated randomly in each episode, so dynamics are learned on noisy data. State components are clipped to $[-1, +1]$ to reduce the effect of outliers, and $\textrm{tanh}$ output activations predict the position and velocity components and $\textrm{sigmoid}$ predicts leg-ground contact. Furthermore, source dynamics are learned offline on data collected during policy training to avoid the moving target problem and improve learning stability. We obtain MSE of only $\sim10^{-3}$ for the land and low-mass source task dynamics, highlighting the difficulty of learning accurate dynamics for ground contact. The test performance averaged over 10 trials is shown in Figure~\ref{fig:lander_learning}. Figure~\ref{fig:mars_lander_trajectories} illustrates the output of the mixture on state trajectories obtained during training. 

\subsection{Discussion}

MARS consistently outperforms all baselines, in terms of sample efficiency and solution quality, and MAPSE outperforms UCB, as shown in Figure~\ref{fig:learning_curve}. Figure~\ref{fig:learning_representation} provide one possible explanation for this, namely the ability of the mixture model to converge to good mixtures even when presented with imperfect source dynamics as in Transfer-SparseLunarLander. Furthermore, on all three domains, MARS achieves asymptotic performance comparable to, or better, than the best single potential function $\Phi1, \Phi2$ etc. Interestingly, although MARS consistently outperforms CAPS, MAPSE only does so on Transfer-SparseLunarLander. This reaffirms our hypothesis in Section~\ref{subsec:mars} that reward shaping can improve generalization on test data with little tuning. Furthermore, we conjecture that the inconsistent performance of CAPS is due to its reliance on fluctuating Q-values, that is mitigated in MARS and MAPSE by their reliance instead on more stable samples of the dynamics. 

\section{Related Work}
\label{sec:related}

Using state-dependent knowledge to contextually reuse multiple source policies is a relatively new topic in transfer learning. \citet{rajendran2015attend} used a soft attention mechanism to learn state-dependent weightings over source tasks, and then transferred either policies or values. \citet{li2018reinforcement} proposed \emph{Two-Level Q-learning}, in which the agent learns to select the most trustworthy source task in each state in addition to the optimal action. The selection of source policies can be seen as an outer optimization problem. The \emph{Context-Aware Policy Reuse} (i.e., CAPS) algorithm of \citet{li2019context} used options to represent selection of source policies as well as target actions, learning Q-values and option termination conditions simultaneously. However, these two papers are limited to critic-based approaches with finite action spaces. To fill this gap, \citet{kurenkov2019ac} proposed \emph{AC-Teach}, which uses Bayesian DDPG to learn probability distributions over Q-values corresponding to student and teacher actions, and Thompson sampling for selecting exploratory actions from them. However, their inference technique is considerably different from ours, and is specific to the actor-critic setting. Our paper complements existing work by using source task dynamics rather than Q-values to reason about task similarity, and is compatible with both model-based and model-free RL.

\emph{Potential-based reward shaping} (PBRS) was first introduced in \citet{ng1999policy} for constructing dense reward signals without changing the optimal policies. Later, \citet{wiewiora2003principled} and \citet{devlin2012dynamic} extended this to action- and time-dependent shaping, respectively. More recently, \citet{harutyunyan2015expressing} combined these two extensions into one framework and used it to incorporate arbitrary reward functions. \citet{brys2015policy} made the connection between PBRS and policy reuse, by turning a single source policy into a binary reward signal and applying \citet{harutyunyan2015expressing}. Later, \citet{suay2016learning} recovered a potential function from demonstrations directly using inverse RL. Our paper extends \citet{brys2015policy} by reusing \emph{multiple} source policies in a \emph{state-dependent} way that is compatible with modern deep RL. Thus, our paper advances the state-of-the-art in policy transfer and contributes to the expanding body of research into reward shaping. 

\section{Conclusion}
\label{sec:conclusion}

We investigated transfer of policies from multiple source tasks with common sub-goals but different dynamics. We showed, theoretically, how errors in dynamics are related to errors in policy values. We then used estimates of source task dynamics to contextually measure similarity between source and target tasks using a deep mixture model. 
We introduced MARS and MAPSE to use this information to improve training in the target task. Experiments showed strong performance of MARS and thus the advantages of leveraging more stable dynamics, as well as reward 
shaping, as a novel means of deep contextual transfer.



\begin{ack}
This work was funded by a DiDi Graduate Student Award. The authors would like to thank Nicolas Carrara for his careful review of the draft and suggestions that significantly improved this paper.
\end{ack}

\small
\bibliographystyle{abbrvnat}
\bibliography{main}

\newpage
\setcounter{page}{1}
\normalsize

\section*{Appendix}

\subsection*{Derivation of Equation~(\ref{eqn:theta_ascent})}

First, since $a_{i}(s; \bm{\theta}) \propto \exp{(z_i^a(s; \bm{\theta}))}$ corresponds to a softmax function, we can apply the chain rule:
\begin{align*}
    \sum_{i=1}^n \hat{P}_i(s'|s,a)\, \nabla_{\bm{\theta}} a_{i}(s; \bm{\theta}) 
    &= \sum_{i=1}^n \hat{P}_i(s'|s,a)\, \sum_{j=1}^n \frac{\partial a_i(s; \bm{\theta})}{\partial z_j^a}\nabla_{\bm{\theta}} z_j^a(s; \bm{\theta}) \\
    &= \sum_{j=1}^n \nabla_{\bm{\theta}} z_j^a(s; \bm{\theta}) \sum_{i=1}^n \hat{P}_i(s'|s,a) \frac{\partial a_i(s; \bm{\theta})}{\partial z_j^a} \\
    &= \sum_{j=1}^n \nabla_{\bm{\theta}} z_j^a(s; \bm{\theta}) \sum_{i=1}^n \hat{P}_i(s'|s,a) a_i(s; \bm{\theta}) \left(\delta_{ij} - a_j(s; \bm{\theta}) \right).
\end{align*}
Then, dividing both sides by the normalizing constant $\sum_{i=1}^n \hat{P}_i(s'|s,a)\, a_{i}(s; \bm{\theta})$, we obtain:
\begin{align*}
    \nabla_{\bm{\theta}} \mathcal{L}(\bm{\theta}) 
    &= -\nabla_{\bm{\theta}} \log\left(\sum_{i=1}^n \hat{P}_i(s'|s,a)\, a_{i}(s; \bm{\theta})\right) 
    = -\frac{\sum_{i=1}^n \hat{P}_i(s'|s,a)\, \nabla_{\bm{\theta}} a_{i}(s; \bm{\theta}) }{\sum_{i=1}^n \hat{P}_i(s'|s,a)\, a_{i}(s; \bm{\theta})} \\
    &= -\sum_{j=1}^n \nabla_{\bm{\theta}} z_j^a(s; \bm{\theta}) \sum_{i=1}^n p_i(s; \bm{\theta}) \left(\delta_{ij} - a_j(s; \bm{\theta}) \right) \\
    &= \sum_{j=1}^n \nabla_{\bm{\theta}} z_j^a(s; \bm{\theta}) \left( a_j(s; \bm{\theta}) - p_j(s; \bm{\theta})\right).
\end{align*}
This is exactly (\ref{eqn:theta_ascent}).

\subsection*{Proof of Theorem~\ref{theorem:main}}

First, observe that for any stochastic matrix $\mathbf{P}$, $\|\mathbf{P}\|=1$, where $\|\cdot \|$ is the infinity norm, and $\mathbf{I} - \gamma \mathbf{P}$ is always invertible, since the eigenvalues of $\gamma \mathbf{P}$ always lie in the interior of the unit circle for $\gamma < 1$. Therefore, 
    \begin{align*}
        \|(\mathbf{I} - \gamma \mathbf{P})^{-1}\| 
        &= \left\|(\gamma \mathbf{P})^0 + (\gamma \mathbf{P})^1 + \dots \right\| \leq \sum_{t=0}^\infty \gamma^t \| \mathbf{P} \|^t = \frac{1}{1 - \gamma}.
    \end{align*}
    To simplify notation, we write $\mathbf{V}_1 = \mathbf{\hat{V}}^{\pi}, \,\mathbf{V}_2 = \mathbf{V}^{\pi}, \,\mathbf{P}_1 = \mathbf{\hat{P}}^{\pi}$ and $\mathbf{P}_2 = \mathbf{{P}}^{\pi}$. Then $\mathbf{V}_1 = (\mathbf{I} - \gamma \mathbf{P}_1)^{-1} \mathbf{R}$ and $\mathbf{V}_2 = (\mathbf{I} - \gamma \mathbf{P}_2)^{-1} \mathbf{R}$. Now, making use of the identity $\mathbf{X}^{-1} - \mathbf{Y}^{-1} = \mathbf{X}^{-1}(\mathbf{Y}-\mathbf{X})\mathbf{Y}^{-1}$, and the triangle inequality, we have
    \begin{align*}
        \| \mathbf{\hat{V}}^{\pi} - \mathbf{V}^{\pi} \| &= \| (\mathbf{I} - \gamma \mathbf{P}_1)^{-1} \mathbf{R} - (\mathbf{I} - \gamma \mathbf{P}_2)^{-1} \mathbf{R} \| \\
        &\leq \| (\mathbf{I} - \gamma \mathbf{P}_1)^{-1}  - (\mathbf{I} - \gamma \mathbf{P}_2)^{-1} \| \|\mathbf{R} \| \\
        &= \| (\mathbf{I} - \gamma \mathbf{P}_1)^{-1} \gamma (\mathbf{P}_2-\mathbf{P}_1) (\mathbf{I} - \gamma \mathbf{P}_2)^{-1} \| \|\mathbf{R} \| \\
        &\leq \gamma  \|(\mathbf{I} - \gamma \mathbf{P}_1)^{-1}\| \| (\mathbf{I} - \gamma \mathbf{P}_2)^{-1}\| \|\mathbf{P}_2-\mathbf{P}_1 \| \|\mathbf{R} \| \\
        &\leq \gamma  \left(\frac{1}{1 - \gamma}\right)^2 \|\mathbf{P}_2-\mathbf{P}_1 \| \|\mathbf{R} \|,
    \end{align*}
    and so the proof is complete. \qedsquare

\subsection*{Pseudocode}

Algorithm~\ref{alg:behavior} illustrates the behavior policy for MAPSE. Algorithm~\ref{alg:main} is the main training loop for MAPSE and MARSE in an episodic reinforcement learning setting.

\begin{algorithm}[!htb]
	\caption{Behavior Policy for MAPSE}
	\label{alg:behavior}
	\algsetup{linenosize=\footnotesize}
        \footnotesize
	\begin{algorithmic}
	    \REQUIRE $s_t$, $\mathbf{a}(s_t; \bm{\theta}_t)$, $p_t$, $\Pi=\lbrace \pi_1^*, \dots \pi_n^*\rbrace$, $\pi^{b}$ \COMMENT{current state, mixture model, current $p_t$, source policy library, target behavior policy}
	    \STATE $\xi_t \sim \textit{Uniform}(0,1)$
	    \IF{$\xi_t < p_t$}
	        \STATE $i_t \sim \mathbf{a}(s_t; \bm{\theta}_t); \, a_t = \pi_{i_t}^*(s_t)$
	    \ELSE
	        \STATE $a_t = \pi^b(s_t)$
	    \ENDIF
	    \STATE Reduce $p_t$
	    \STATE \textbf{return} $a_t$
	\end{algorithmic}
\end{algorithm}

\begin{algorithm}[!htb]
	\caption{Model-Aware Policy Reuse (MAPSE, MARS)}
	\label{alg:main}
	\algsetup{linenosize=\footnotesize}
        \footnotesize
	\begin{algorithmic}
	    \REQUIRE $\hat{P}_{i}(s'|s,a)$ (or $\hat{f}_{i}(s,a)$ and $\rho_i$) for $i = 1,2 \dots n$, $\Pi=\lbrace \pi_1^*,\dots \pi_n^*\rbrace$, $\pi$, $\pi^b$, $\bm{\theta}$, $\lambda_m$, $\mathcal{D}$, (optionally $p_t$ or $c$) \COMMENT{learned source dynamics, source policy library, learned target policy, behavior policy, mixture model weights, learning rate, replay buffer}
		\FOR{episode $m=1, 2, \dots$}
		    \STATE Sample an episode $(s_0,a_0,r_0\dots s_T)$ from the environment, where $a_t$ is defined by Algorithm~\ref{alg:behavior} (MAPSE) or $r_t$ is defined by (\ref{eqn:rs}) and (\ref{eqn:potential}) (MARS)
		    \STATE Store $(s_0,a_0,r_0\dots s_T)$ in $\mathcal{D}$
		    \STATE Train $\pi$ on random mini-batches sampled from $\mathcal{D}$
		    \STATE Update $\bm{\theta}$ using gradient descent (\ref{eqn:theta_ascent}) on random mini-batches sampled from $\mathcal{D}$, e.g. 
		    \[
		        \bm{\theta} = \bm{\theta} - \lambda_m \nabla_{\bm{\theta}} \mathcal{L}(\bm{\theta})
		    \]
		\ENDFOR
		\STATE \textbf{return} $\pi$
	\end{algorithmic}
\end{algorithm}

\subsection*{Visual Illustration of the Transfer-Maze Domain}

\begin{figure}[!htb]
    \centering
    \begin{subfigure}{0.8\textwidth}
        \begin{tabular}[b]{cccc}
             \includegraphics[width=0.2\linewidth]{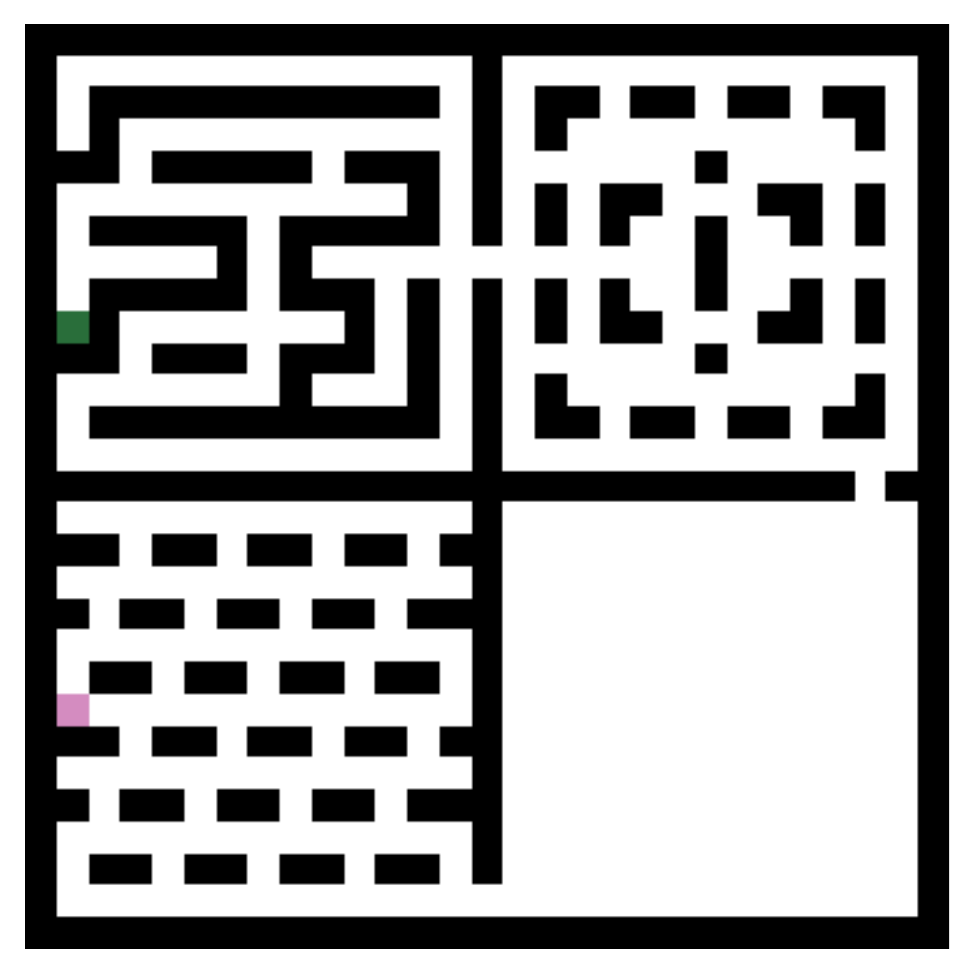} & \includegraphics[width=0.2\linewidth]{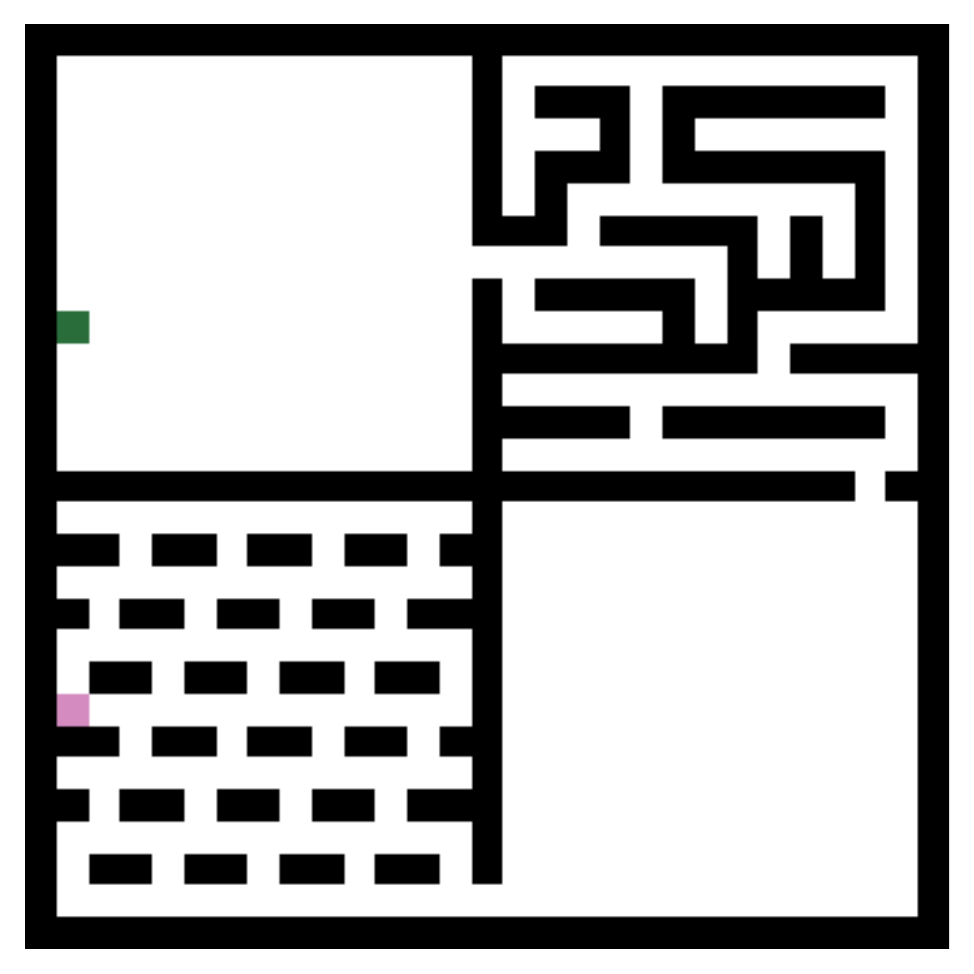} & \includegraphics[width=0.2\linewidth]{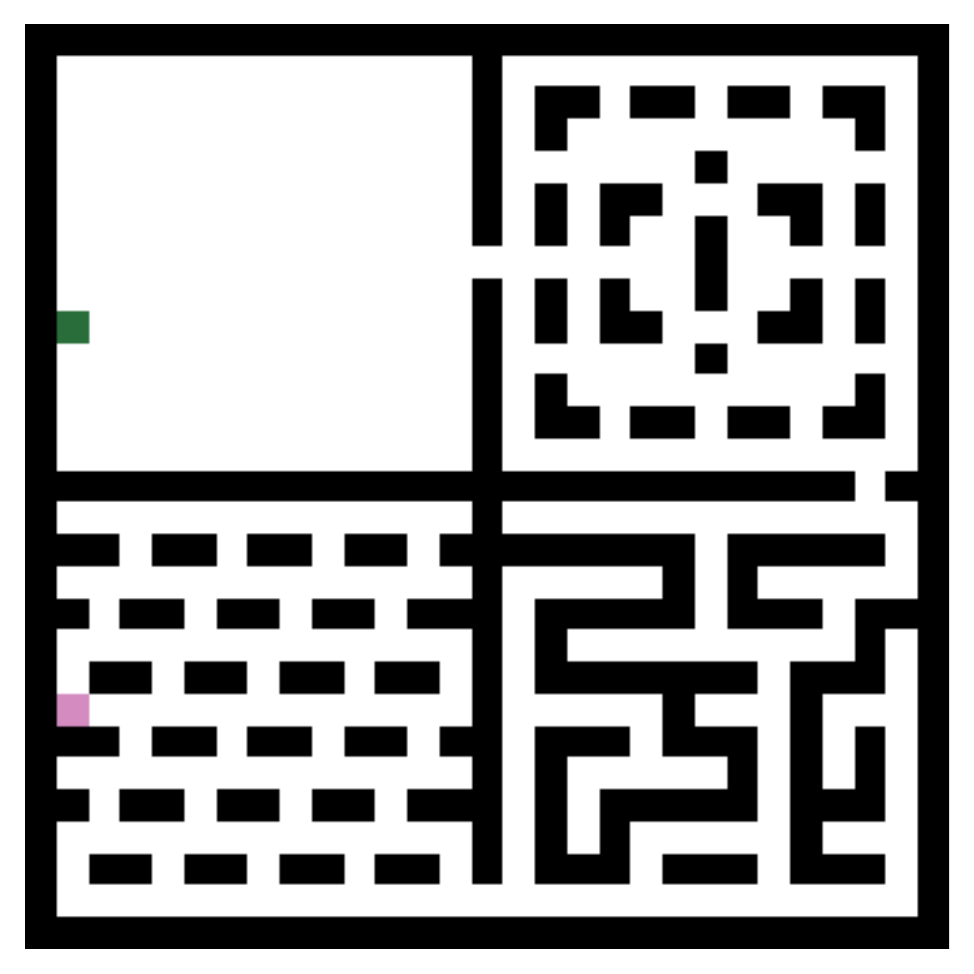} & \includegraphics[width=0.2\linewidth]{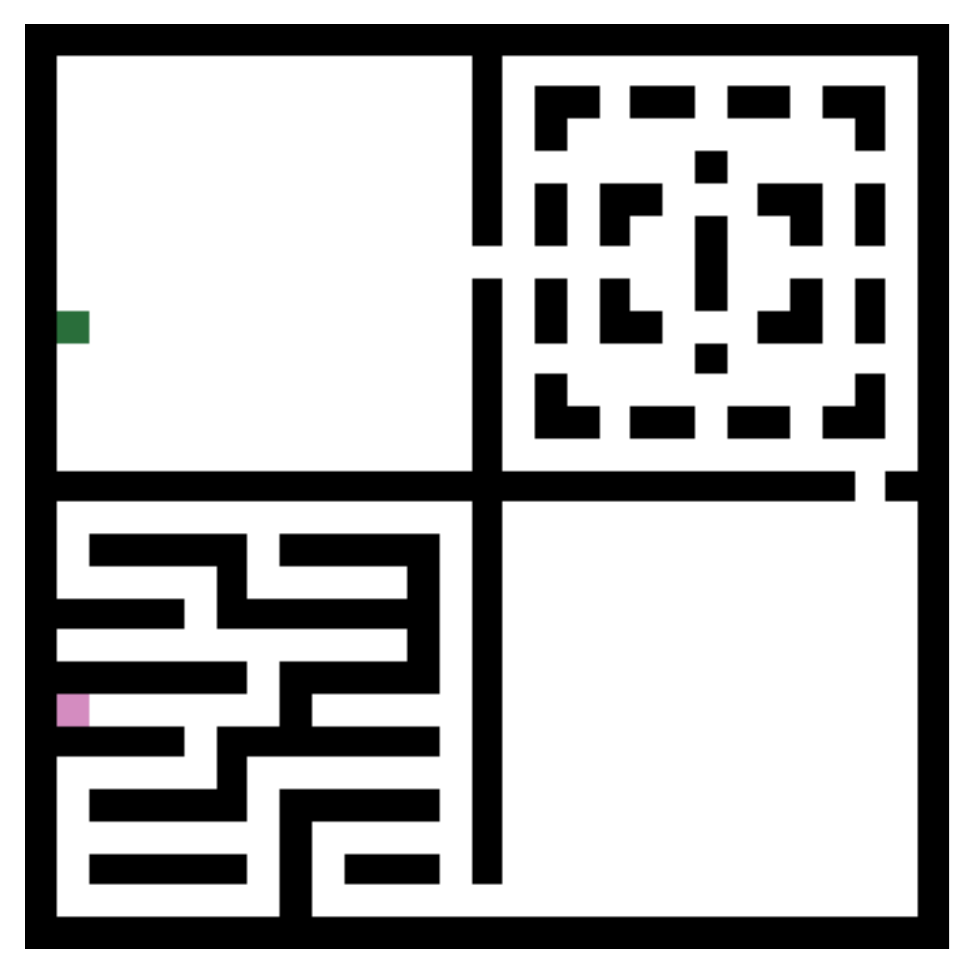}
        \end{tabular}
        \caption{Source tasks.}
    \end{subfigure}\hfill%
    \begin{subfigure}{0.195\textwidth}
        \centering
            \includegraphics[width=0.955\linewidth]{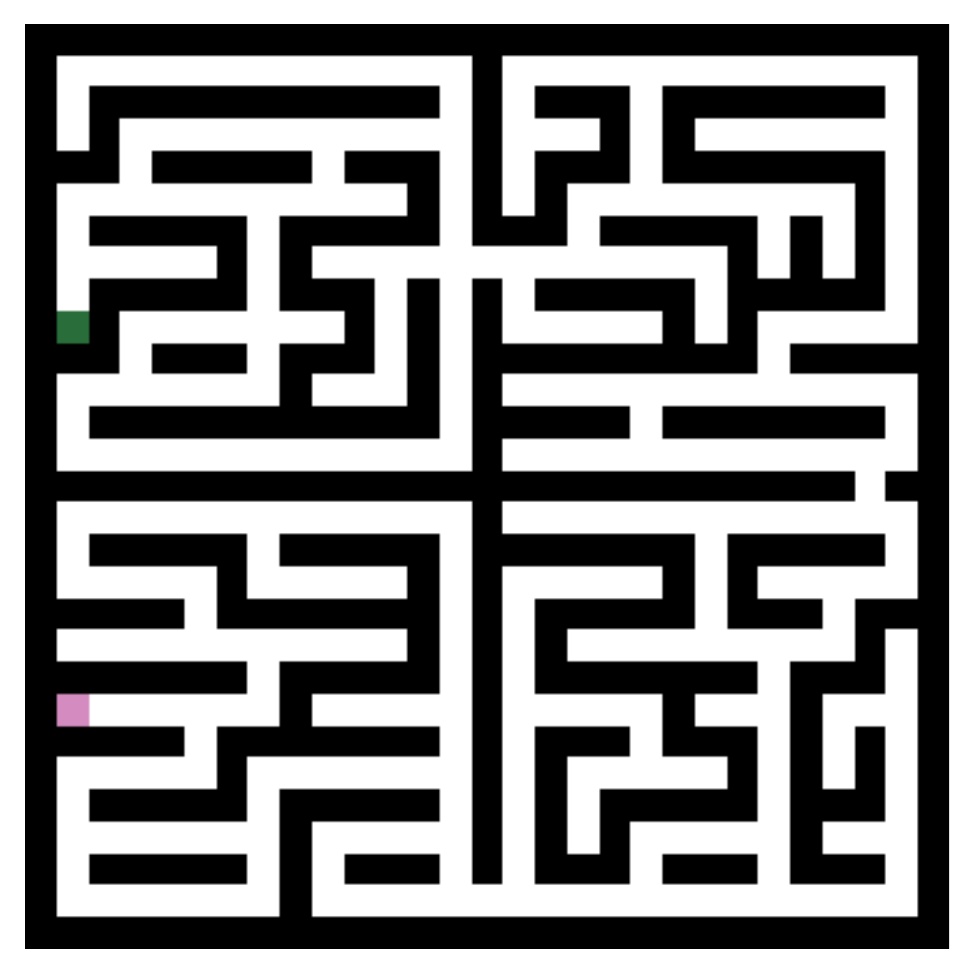}
            \caption{Target tasks.} 
    \end{subfigure}
        \caption{The Transfer-Maze environment. The agent always begins in a fixed cell (highlighted in green). The goal of the agent is to navigate to the fixed target cell (highlighted in red). (a) Four source tasks with different arrangement of obstacles in each room. (b) Target task consists of four rooms with different arrangements of obstacles in each room.}
        \label{fig:maze}
\end{figure}

Figure~\ref{fig:maze} provides a visual description of the Transfer-Maze domain.

\subsection*{Additional Plots}

\begin{figure}[!htb]
    \centering
        \begin{subfigure}[tb]{0.3\textwidth}
        \centering
            \includegraphics[width=0.95\linewidth]{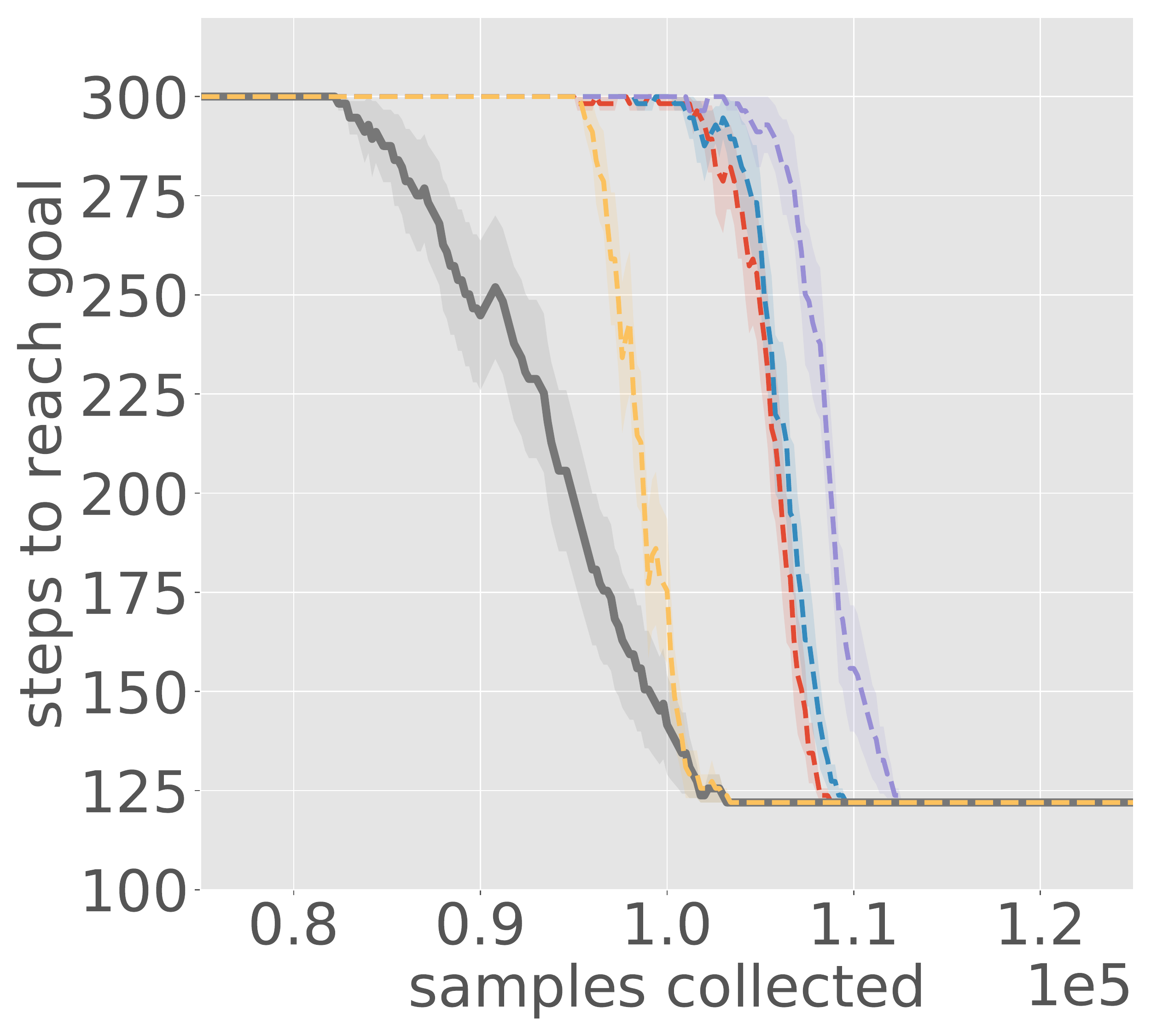}
            \caption{Transfer-Maze}
            \label{fig:maze_maps_learning}
        \end{subfigure}\hfill
        \begin{subfigure}[tb]{0.3\textwidth}
        \centering
            \includegraphics[width=0.95\linewidth]{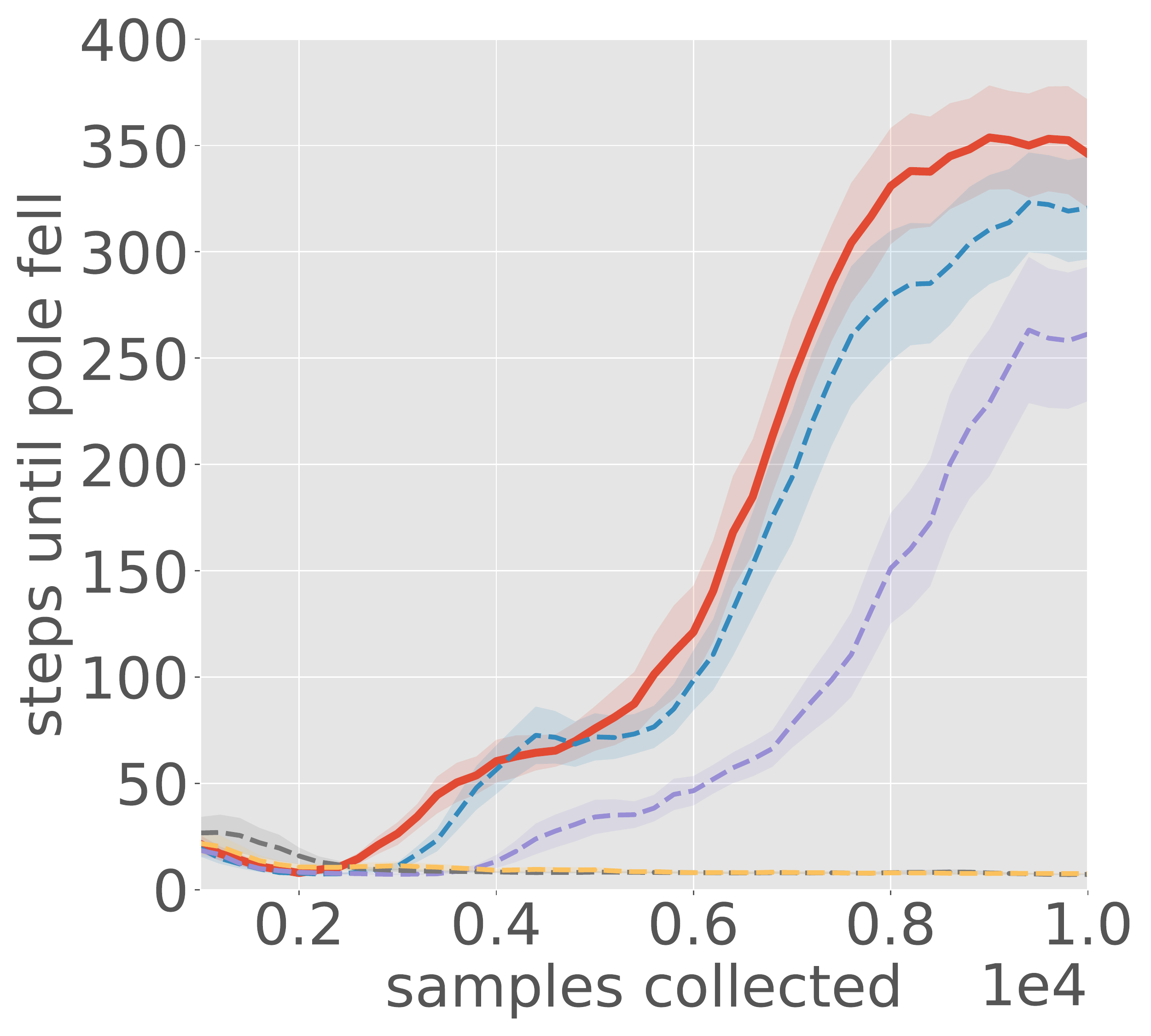}
            \caption{Transfer-CartPole}
            \label{fig:cartpole_maps_learning}
        \end{subfigure}\hfill
        \begin{subfigure}[tb]{0.3\textwidth}
            \centering
            \includegraphics[width=0.95\linewidth]{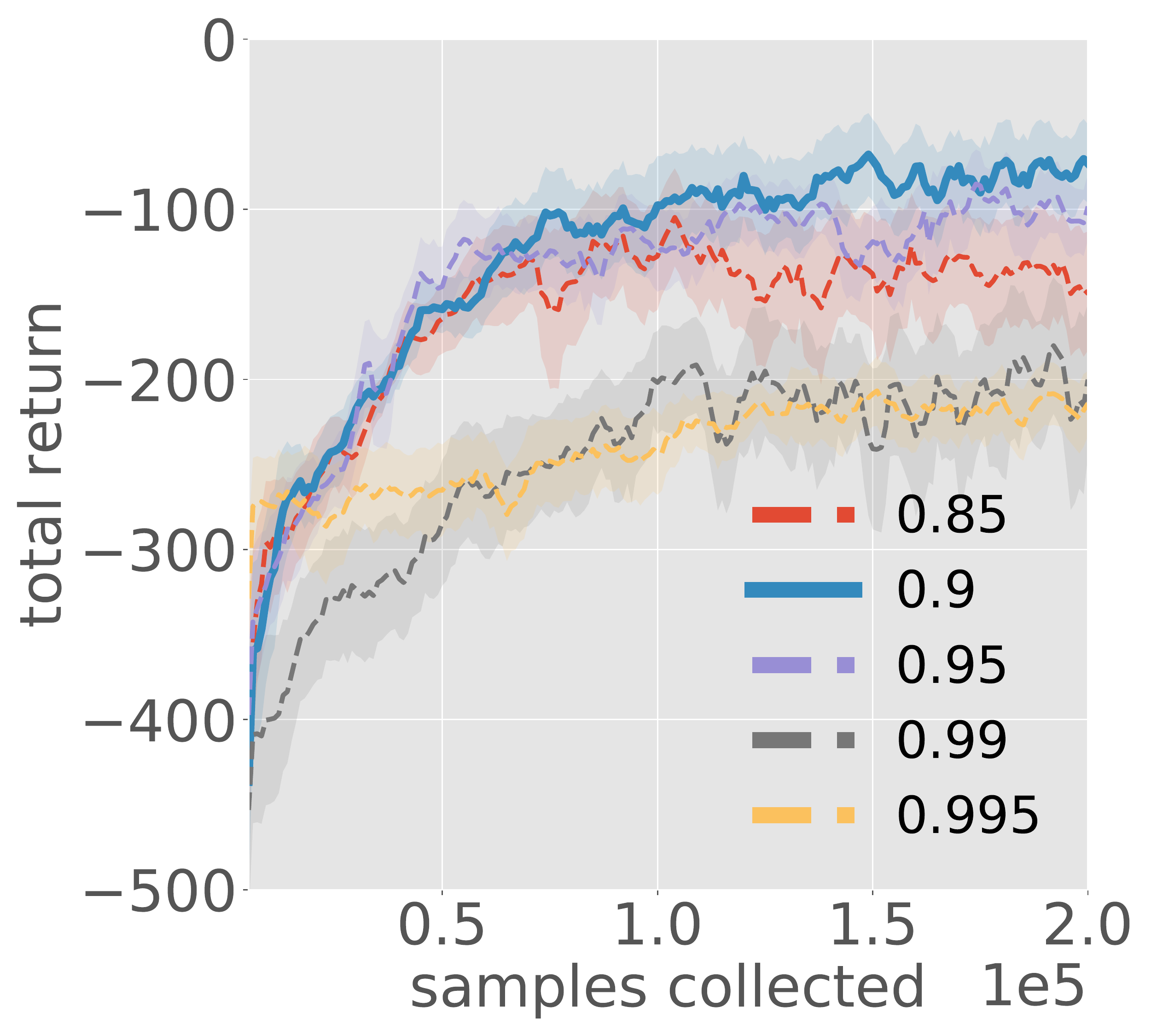}
            \caption{Tr.-SparseLunarLander}
            \label{fig:lander_maps_learning}
        \end{subfigure}
        \caption{Smoothed mean test performance using the greedy policy for different value of $p_t$ for MAPSE: (a) number of steps to reach goal (b) number of steps balanced (c) total return and standard error over number of training samples collected. We ran 20 trials for Transfer-Maze and Transfer-CartPole and 10 trials for Transfer-SparseLunarLander.}
        \label{fig:maps_learning_curve}
\end{figure}

\begin{figure}[!htb]
    \centering
        \begin{subfigure}[tb]{0.3\textwidth}
        \centering
            \includegraphics[width=0.95\linewidth]{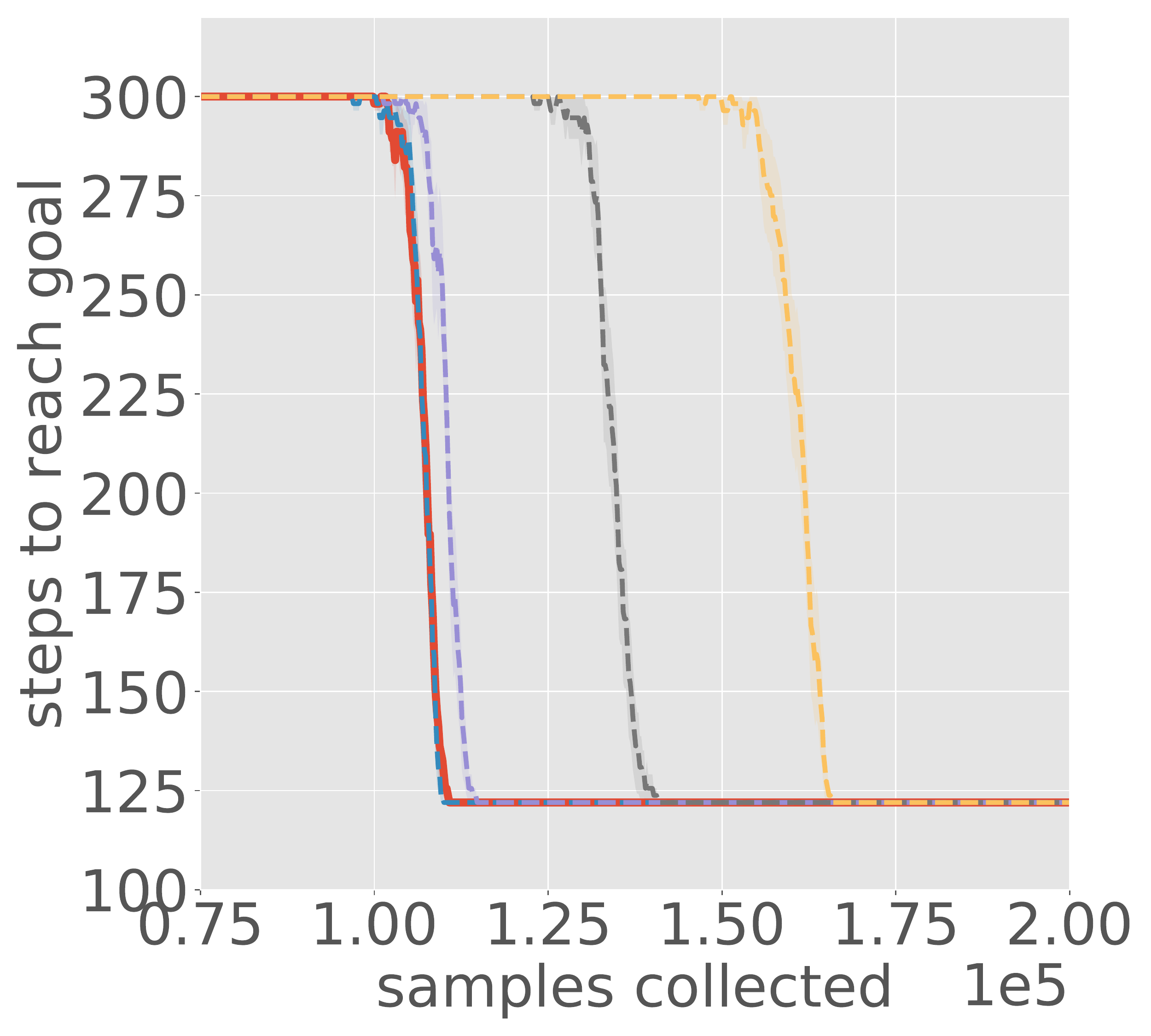}
            \caption{Transfer-Maze}
            \label{fig:maze_ucb_learning}
        \end{subfigure}\hfill
        \begin{subfigure}[tb]{0.3\textwidth}
        \centering
            \includegraphics[width=0.95\linewidth]{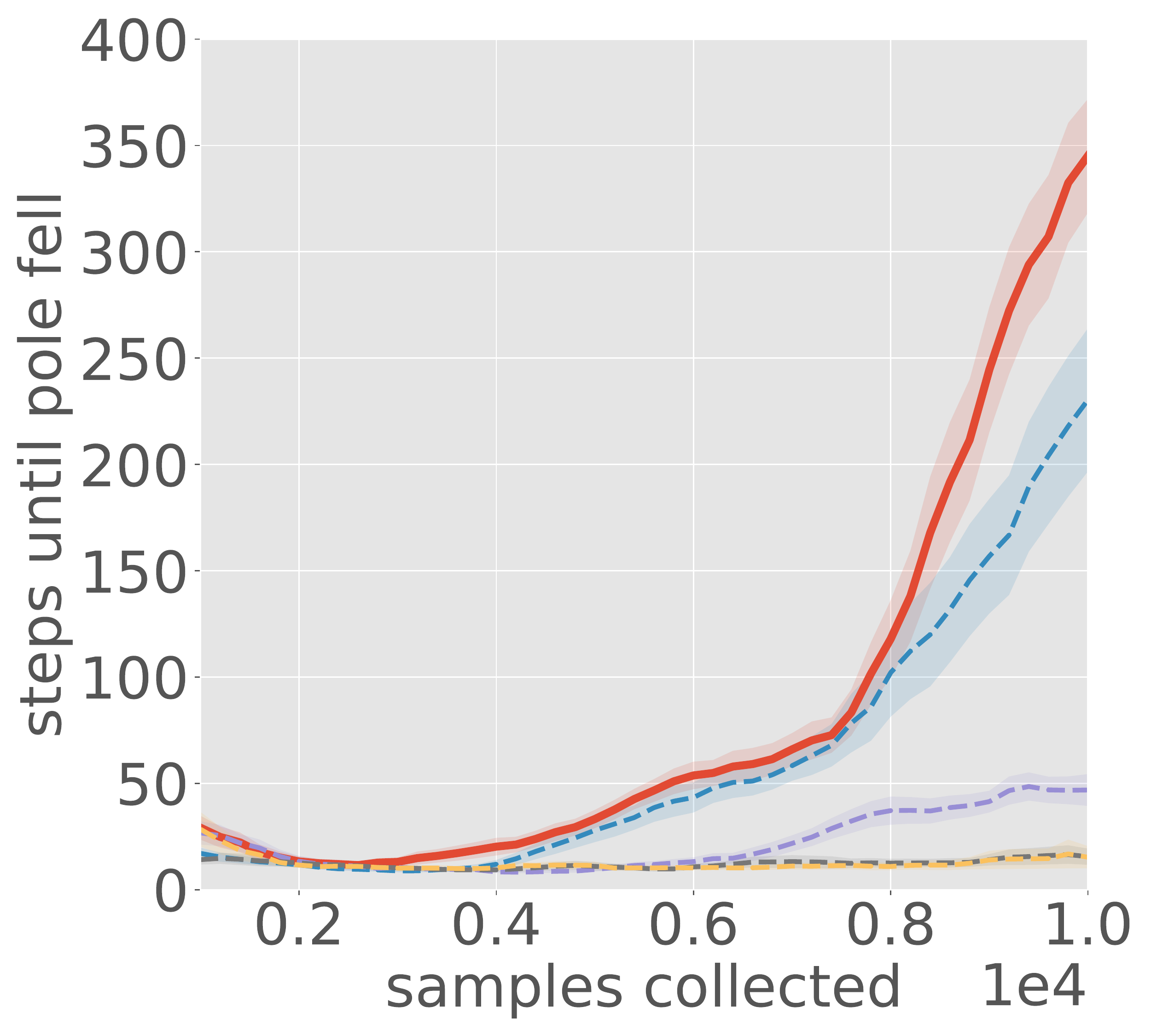}
            \caption{Transfer-CartPole}
            \label{fig:cartpole_ucb_learning}
        \end{subfigure}\hfill
        \begin{subfigure}[tb]{0.3\textwidth}
            \centering
            \includegraphics[width=0.95\linewidth]{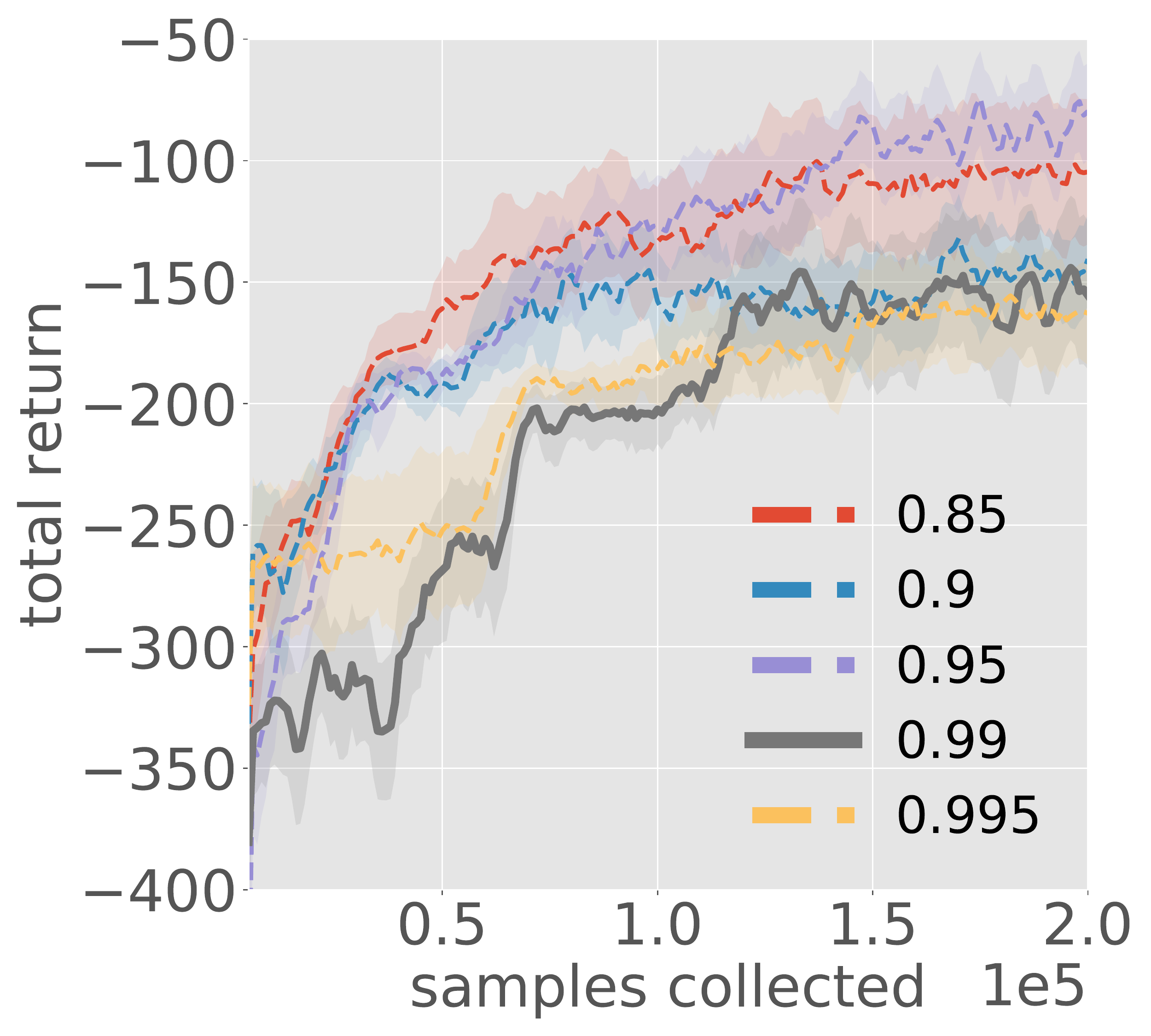}
            \caption{Tr.-SparseLunarLander}
            \label{fig:lander_ucb_learning}
        \end{subfigure}
        \caption{Smoothed mean test performance using the greedy policy for different value of $p_t$ for UCB: (a) number of steps to reach goal (b) number of steps balanced (c) total return and standard error over number of training samples collected. We ran 20 trials for Transfer-Maze and Transfer-CartPole and 10 trials for Transfer-SparseLunarLander.}
        \label{fig:ucb_learning_curve}
\end{figure}

\begin{figure}[!htb]
    \centering
        \begin{subfigure}[tb]{0.3\textwidth}
        \centering
            \includegraphics[width=0.95\linewidth]{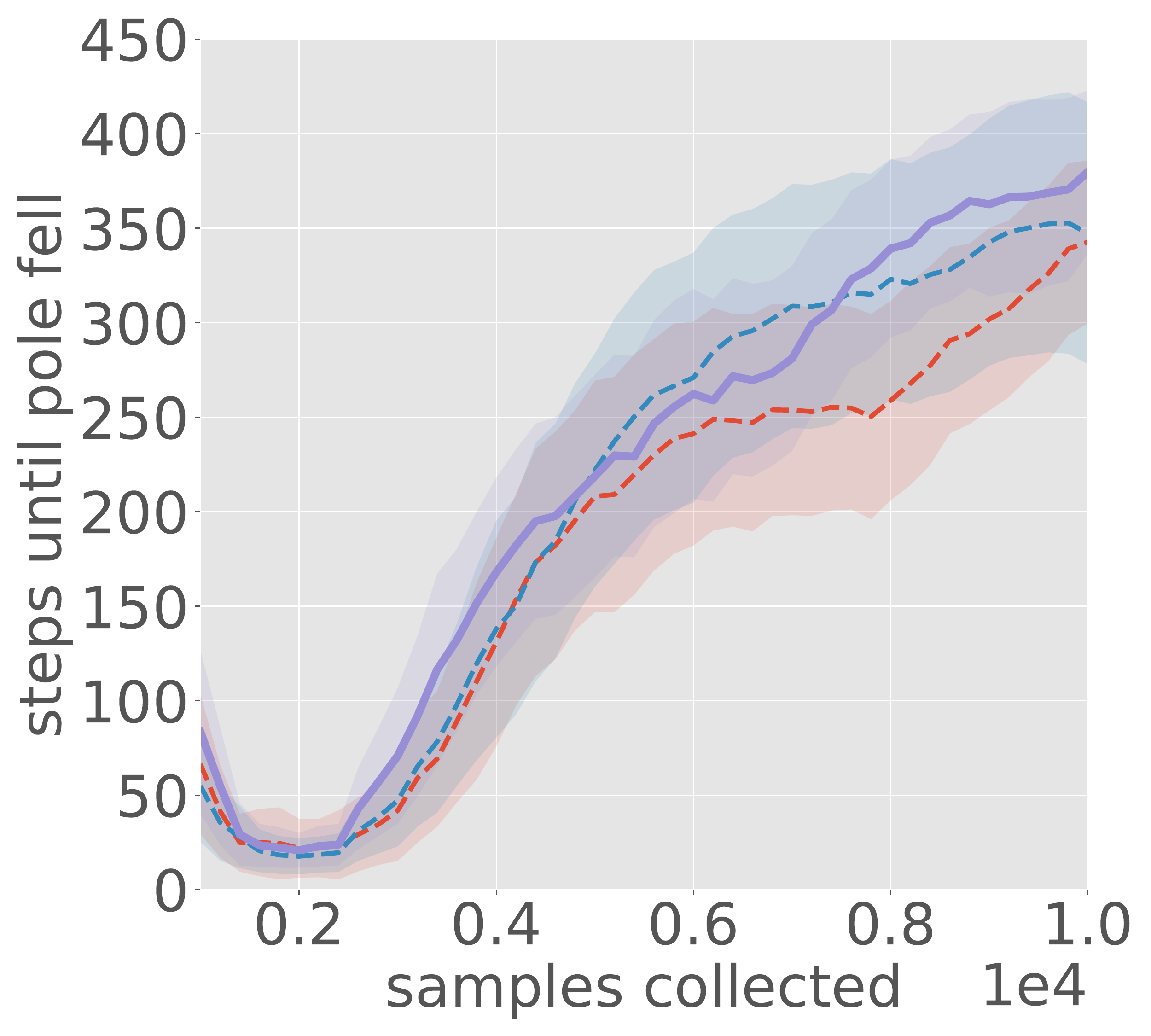}
            \caption{Transfer-CartPole}
            \label{fig:cartpole_caps_learning}
        \end{subfigure}
        \begin{subfigure}[tb]{0.3\textwidth}
        \centering
            \includegraphics[width=0.95\linewidth]{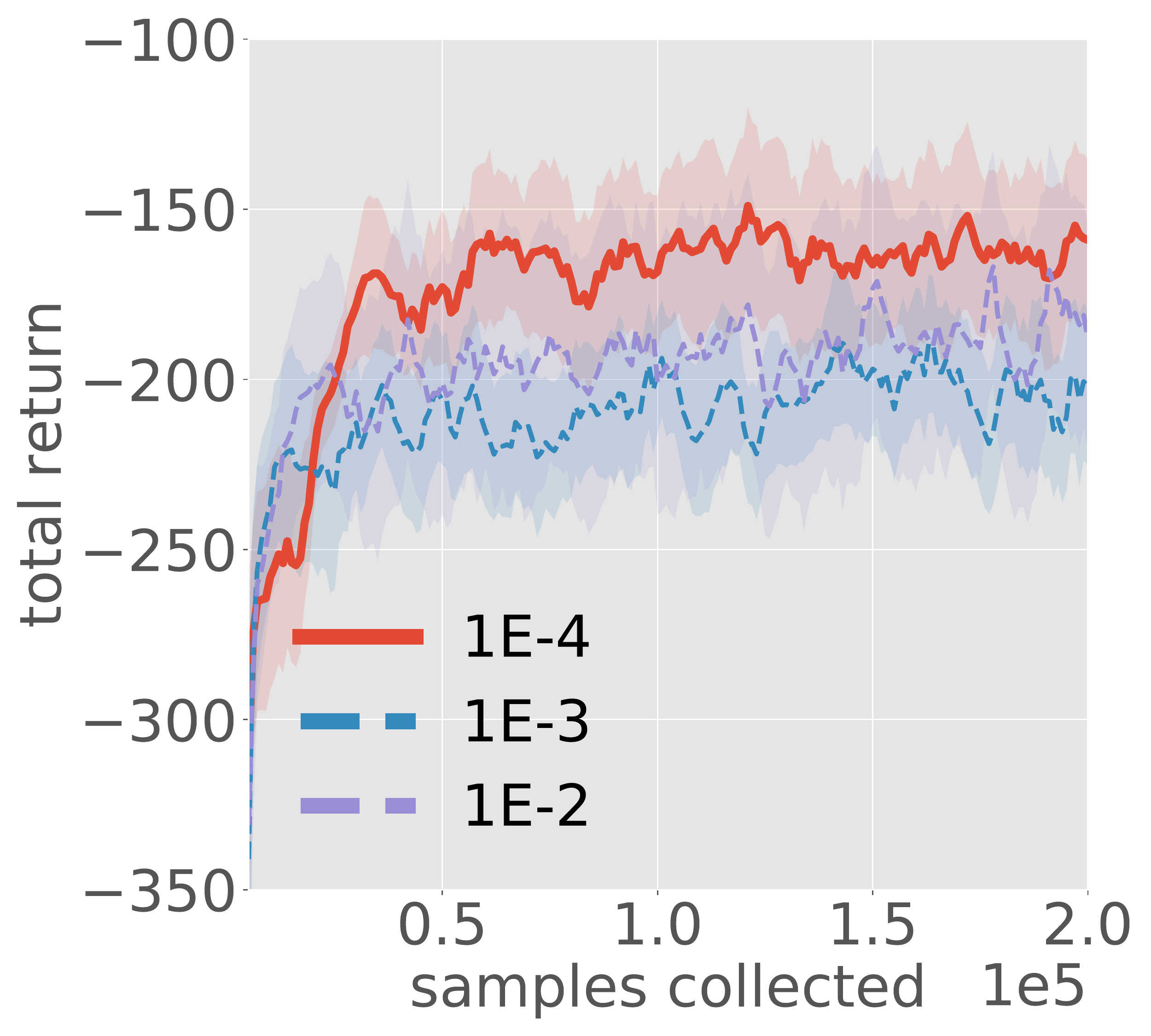}
            \caption{Tr.-SparseLunarLander}
            \label{fig:lander_caps_learning}
        \end{subfigure}
        \caption{Smoothed mean test performance using the greedy policy for different value of termination learning rate for CAPS: (a) number of steps balanced (c) total return and standard error over number of training samples collected. We ran 20 trials for Transfer-CartPole and 10 trials for Transfer-SparseLunarLander.}
        \label{fig:caps_learning_curve}
\end{figure}

Figure~\ref{fig:maps_learning_curve} illustrates the test performance on all three transfer learning experiments (Transfer-Maze, Transfer-CartPole and Transfer-SparseLunarLander) using different values of the reuse parameter $p_t$ for the MAPSE algorithm. Figure~\ref{fig:ucb_learning_curve} demonstrates the test performance on all transfer learning experiments using different values of the reuse parameter $p_t$ for the UCB-based policy reuse algorithm~\citep{li2018optimal}. Figure~\ref{fig:caps_learning_curve} illustrates the test performance on all transfer learning experiments using different learning rates $\beta$ for the last layer of the option-value network for the option-based context-aware policy reuse algorithm CAPS~\citep{li2019context}. 

\subsection*{Further Implementation Details and Choice of Hyperparameters}

All code was written and executed using Eclipse PyDev running Python 3.7. All neural networks were initialized and trained using Keras with TensorFlow backend (version 1.14), and weights were initialized using the default setting. The Adam optimizer was used to train all neural networks. Experiments were run on an Intel 6700-HQ Quad-Core processor with 8 GB RAM running on the Windows 10 operating system. The hyper-parameter settings used in the experiments are listed in Table~\ref{table:hyper}. 

\begin{table}[!htb]
\resizebox{\textwidth}{!}{
    \begin{tabular}{lllll}
		\toprule
		\multicolumn{2}{c}{Parameters}                   \\
		\cmidrule(r){1-2}
		Name     & Description     & Transfer-Maze & Transfer-CartPole & Transfer-SparseLunarLander \\
		\midrule
		$T$ & maximum roll-out length & 300 & 500 & 1000 \\
    $\gamma$ & discount factor & 0.95 & 0.98 & 0.99 \\
    $\varepsilon_t$ & exploration probability & 0.12 & $\max\lbrace 0.01, 0.99^t \rbrace$ & $\max\lbrace 0.01, 0.9925^t \rbrace$ \\
    & learning rate of Q-learning* & 0.08, 0.8 & & \\
    & replay buffer capacity & & 5000 & 20000 \\
    $B$ & batch size & & 32 & 64 \\
    $Q$ & topology of DQN & & 4-40-40-4 & 8-120-100-4 \\
    & hidden activation of DQN & & ReLU & ReLU \\
    & learning rate of DQN* & & 0.0002, 0.0005 & 0.0002, 0.0005 \\
    & learning rate for termination function weights** & 0.4 & 0.01 & 0.0001 \\
    & target network update frequency (in batches) & & 500 & 100 \\
    & L2 penalty of DQN & & $10^{-6}$ & $10^{-6}$ \\
    $\hat{f}_i$ & topology of dynamics model & & 8-50-50-4 & 12-100-100-8 \\
    & hidden activation of dynamics model & & ReLU & ReLU \\ 
    & learning rate of dynamics model & & 0.001 & 0.001 \\ 
    & L2 penalty of dynamics model & & $10^{-6}$ & $10^{-6}$ \\ 
    $\nu_i$ & Gaussian kernel precision & & $5 \times 10^5$ & $5 \times 10^5$ \\ 
    $\mathbf{a}$ & topology of mixture model & 58-30-30-4 & 4-30-30-3 & 8-30-30-3 \\ 
    & hidden activation of mixture & ReLU & ReLU & ReLU \\ 
    $\lambda_{m}$ & learning rate of mixture & 0.001 & 0.001 & 0.001 \\ 
    & training epochs/batch for mixture & 4 & 3 & 1 \\ 
    $c$ & PBRS scaling factor & 1.0 & 2.0 & 20.0 \\ 
    $p_t$ & probability of following source policies*** & $0.99^t$ (MAPSE), $0.85^t$ (UCB) & $0.85^t$ (MAPSE), $0.85^t$ (UCB) & $0.9^t$ (MAPSE), $0.95^t$ (UCB)\\
		\bottomrule
	\end{tabular}}
	\caption{Hyper-parameter settings used in all transfer learning experiments.}
	\label{table:hyper}
\end{table}

\footnotesize
\noindent* we had to decrease the learning rate for MARS and reward shaping using a single policy to avoid instability \\
\noindent** we report the best value found in $\lbrace 0.01, 0.001, 0.0001 \rbrace$ in the deep learning case \\
\noindent*** $p_t = p^t$ where $t$ is the episode number; we report the best $p \in \lbrace 0.85, 0.9, 0.95, 0.99, 0.995 \rbrace$
\normalsize

\end{document}